\pdfoutput=1
\documentclass[lettersize,journal]{IEEEtran}
\usepackage{amsmath,amsfonts}
\usepackage{graphicx}
\usepackage{caption}
\usepackage{subcaption}
\usepackage{amsmath}
\usepackage{amssymb}
\usepackage{booktabs}
\usepackage{bbm}
\usepackage{multirow}
\usepackage{algorithmic}
\usepackage[ruled,linesnumbered]{algorithm2e}
\usepackage{float}
\usepackage[pagebackref,breaklinks,colorlinks]{hyperref}
\usepackage[capitalize]{cleveref}
\crefname{section}{Sec.}{Secs.}
\Crefname{section}{Section}{Sections}
\Crefname{table}{Table}{Tables}
\crefname{table}{Tab.}{Tabs.}

\def\BibTeX{{\rm B\kern-.05em{\sc i\kern-.025em b}\kern-.08em
    T\kern-.1667em\lower.7ex\hbox{E}\kern-.125emX}}
\usepackage{balance}
\begin{document}
\title{Imbalanced Open Set Domain Adaptation via Moving-threshold Estimation and Gradual Alignment}
\author{Jinghan Ru, Jun Tian, Zhekai Du, Chengwei Xiao, Jingjing Li and Heng Tao Shen, {\textit{Fellow, IEEE}}
\thanks{J. Ru, J. Tian, Z.Du, C. Xiao, J. Li and H. T. Shen are with the School of Computer Science and Engineering, University of Electronic Science and Technology of China, Chengdu 611731, China. (Jingjing Li is the corresponding author, e-mail: lijin117@yeah.net).}}

\markboth{Journal of \LaTeX\ Class Files,~Vol.~18, No.~9, September~2020}%
{Imbalanced Open Set Domain Adaptation via Moving-threshold Estimation and Gradual Alignment}

\maketitle

\begin{abstract}
Multimedia applications are often associated with cross-domain knowledge transfer, where Unsupervised Domain Adaptation (UDA) can be used to reduce the domain shifts. Open Set Domain Adaptation (OSDA) aims to transfer knowledge from a well-labeled source domain to an unlabeled target domain under the assumption that the target domain contains unknown classes. Existing OSDA methods consistently lay stress on the covariate shift, ignoring the potential label shift problem. The performance of OSDA methods degrades drastically under intra-domain class imbalance and inter-domain label shift. However, little attention has been paid to this issue in the community. In this paper, the Imbalanced Open Set Domain Adaptation (IOSDA) is explored where the covariate shift, label shift and category mismatch exist simultaneously. To alleviate the negative effects raised by label shift in OSDA, we propose Open-set Moving-threshold Estimation and Gradual Alignment (OMEGA) - a novel architecture that improves existing OSDA methods on class-imbalanced data. Specifically, a novel unknown-aware target clustering scheme is proposed to form tight clusters in the target domain to reduce the negative effects of label shift and intra-domain class imbalance. Furthermore, moving-threshold estimation is designed to generate specific thresholds for each target sample rather than using one for all. Extensive experiments on IOSDA, OSDA and OPDA benchmarks  demonstrate that our method could significantly outperform existing state-of-the-arts. Code and data are available at \href{https://github.com/mendicant04/OMEGA}{https://github.com/mendicant04/OMEGA}. 
\end{abstract}


\section{INTRODUCTION}
\IEEEPARstart{D}{eep} neural networks have shown significant performance on many computer vision tasks, such as image recognition \cite{alexnet, resnet} and semantic segmentation \cite{maskrcnn}. However, common deep learning algorithms rely heavily on the assumption that the training and testing samples follow the same distribution. In many multimedia applications\cite{mma, mmda1, mmda2}, this assumption is not realistic. \textit{Unsupervised domain adaptation} (UDA)\cite{cdcl, uda, udar} aims to transfer a model trained on a labeled source domain to an unlabeled target domain that conforms to a different distribution. Although many UDA methods have achieved great progress by narrowing the discrepancy between the source distribution $p$ and the target distribution $q$ via minimizing a predefined distance metric \cite{mmd,wasser1,dis1} or adversarial alignment \cite{adda, adv2, unif}, most of them rely on the assumption that the label distributions of the two domains are identical. Denoting the input image features as $x$ and the corresponding labels as $y$, vanilla UDA methods only assume the existence of covariate shift (\textit{i.e.}, $p(x) \neq q(x)$ and $p(y|x) = q(y|x)$) but no label shift ($p(y)=q(y)$). 
\begin{figure}
  \begin{subfigure}{\columnwidth}
      \includegraphics[width=\textwidth, trim={0 15mm 0 15mm}, clip]{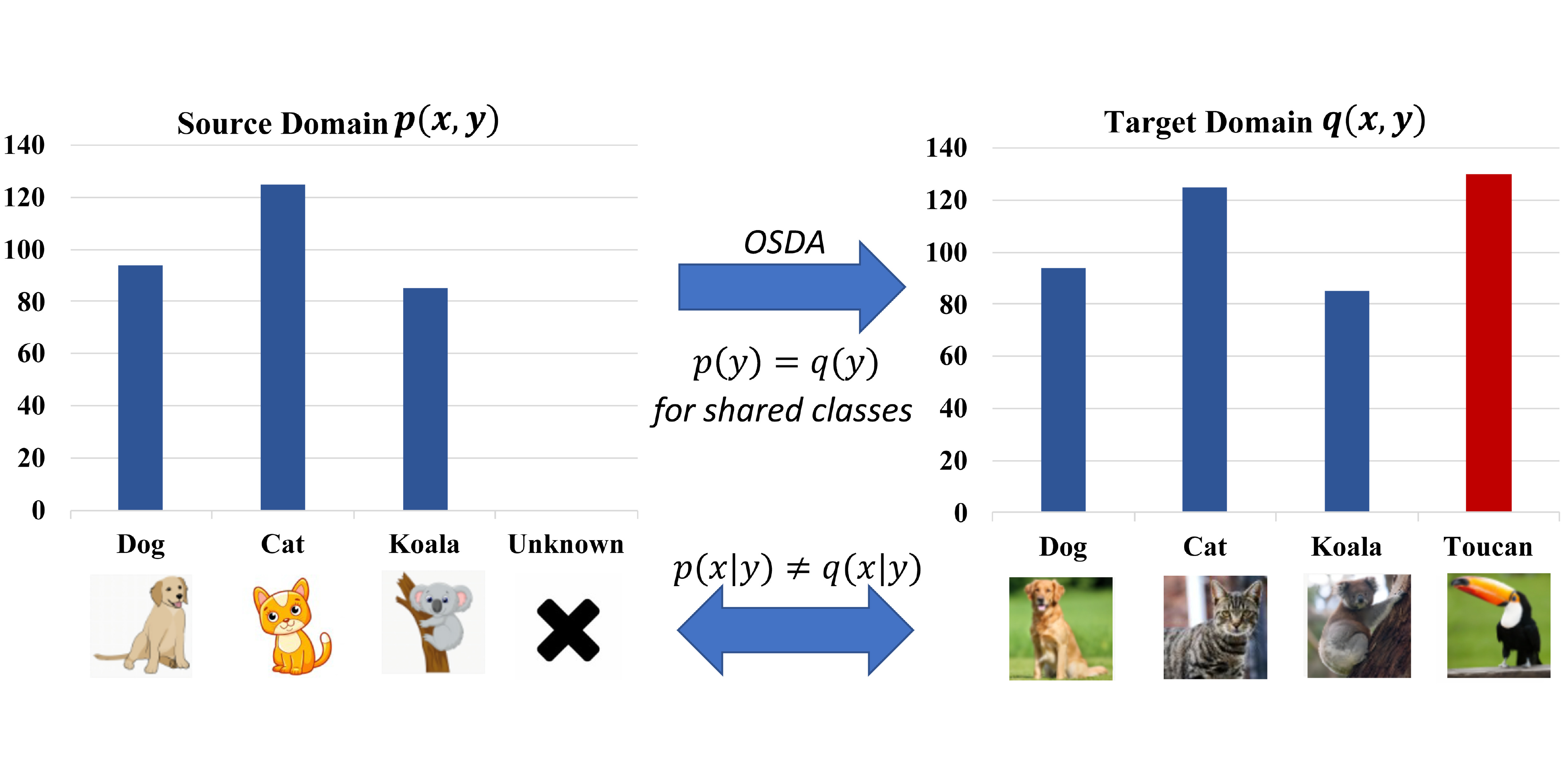}
  \end{subfigure}
  \begin{subfigure}{\columnwidth}
      \includegraphics[width=\textwidth, trim={0 10mm 0 20mm}, clip]{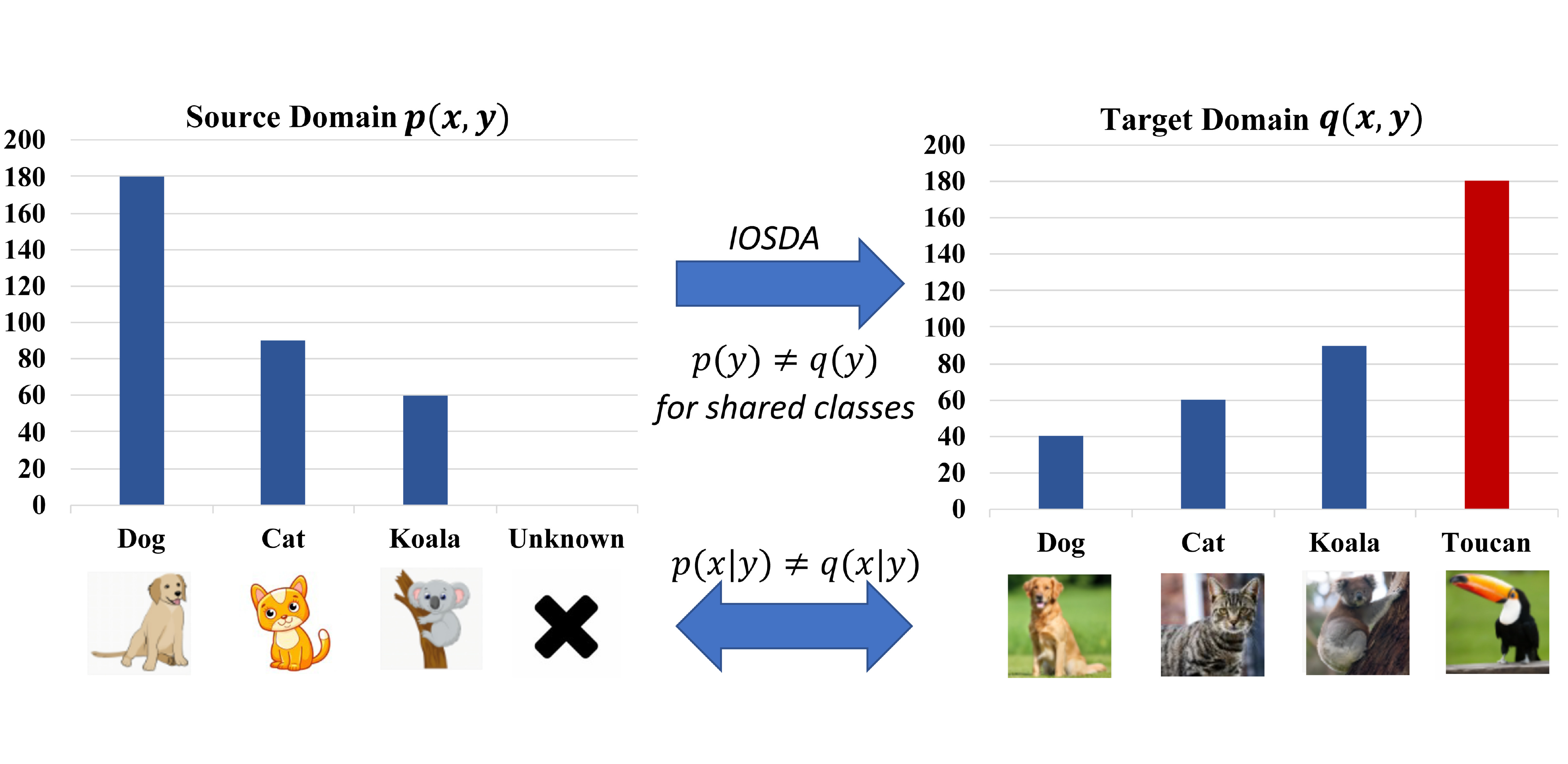}
  \end{subfigure}
  \vspace{-5mm}
  \caption{ {\textbf{Top}}: Standard open set domain adaptation assumes that the label distributions of shared classes are identical, and only covariate shift exists. {\textbf{Bottom}}: Imbalanced
  open set domain adaptation considers a more realistic situation where there exists label shift and severe class imbalance problem simultaneously. For example, suppose we wish to deploy an animal identification system, the training data crawled from the Internet may include more common animals like cats and dogs. However, the users will not identify them as they already know what they are. Instead, they will identify rare animals such as toucans.}
  \label{Fig11}
  \vspace{-4mm}
\end{figure}

However, this assumption can hardly hold true in real-world scenarios. For instance, the trained model may encounter ``unknown'' samples for which $p(y)=0$ in the wild. In this case, those samples from unknown classes will be incorrectly assigned to known classes. To tackle this problem, \textit{Open Set Domain Adaptation} (OSDA) \cite{osda, osbp} has been proposed by enabling the model to recognize those samples that do not belong to the label set of the source domain as ``unknown'' while correctly classifying the other samples in shared classes. Existing OSDA methods typically assume that the prior label distributions $p(y)$ and $q(y)$ for shared classes are identical. However, label shift and intra-domain class imbalance are common problems in practice \cite{bbse}. A well-known example is the long-tailed learning problem \cite{pareto, logit} which occurs in many realistic scenarios such as face recognition \cite{facelt, face1}. Previous works have demonstrated that such class imbalance problem can severely degrade the model performance \cite{coal}. Therefore, we argue that a robust model must be able to handle covariate shift and label shift while correctly recognizing unknown samples simultaneously. 

To tame the aforementioned problem, we take one step beyond conventional OSDA by exploring {\textit{\textbf{Imbalanced Open Set Domain Adaptation}}} (IOSDA), a more realistic but challenging setting as shown in \cref{Fig11}. Note that this could also be seen as an extension of {\textit{Universal Domain Adaptation}} \cite{unida} ({\textit{i.e.}}, $q(y)=0$ for $y$ in source-only classes). The main challenges of IOSDA are two-fold: (1) how to effectively distinguish unknown samples while correctly classifying known samples with a biased classifier; (2) how to learn discriminative and robust features in the presence of covariate shift, label shift and intra-domain class imbalance. To surmount these obstacles, we propose {\textit{Open-set Moving-threshold Estimation and Gradual Alignment (OMEGA)}} framework. Specifically, to overcome the first difficulty, we design a novel schema called moving-threshold estimation to obtain sample-wise thresholds for recognizing unknown samples. Previous OSDA methods mainly treat unknown samples as a whole, ignoring the intrinsic structure within the unknown samples. Based on this observation, we exploit the semantic relationships between different unknown samples and the source domain. Samples that are closer to the source domain will get smaller thresholds and vice versa. As for the second challenge, we propose a novel ``unknown-aware target clustering loss'' to further form tight clusters and clear decision boundaries in the target domain in a self-supervised way. Concretely, we bring the target domain samples with the same pseudo-label closer to each other in the feature space. For unknown samples, they are first divided into different semantic clusters, and only samples within the same cluster are aligned. Moreover, the confidence score is generated for each pair to avoid overconfident predictions. Finally, instead of explicitly aligning the source and target domain which may lead to negative transfer \cite{neg_trans} under large domain gap, we adopt neighborhood clustering loss\cite{dance} and domain-specific batch normalization \cite{dsbb} to achieve gradual alignment. Our contributions can be summarized as follows:
\begin{itemize}
  \item{We extend traditional OSDA to imbalanced open set domain adaptation, which is practical in real-world applications. We propose a novel Open-set Moving-threshold Estimation and Gradual Alignment (OMEGA) framework to tackle it.}
  \item{We propose a novel unknown-aware target clustering loss to form discriminative clusters in the target domain that mitigates the effect of class imbalance and a novel moving-threshold estimation method to obtain sample-wise thresholds for unknown samples.}
  \item{Extensive experiments on three imbalanced benchmark datasets verify that OMEGA outperforms existing OSDA methods in terms of HOS score. The code and data of our method are released, which we believe could help future researchers move towards more practical domain adaptation.}
\end{itemize}

\section{Related Works}
\label{re}

\textbf{Domain Adaptation Under Covariate Shift}. Standard unsupervised domain adaptation assumes that the source and target domain have the same label set and transfers knowledge from the source domain to the target domain under covariate shift \cite{yq}. Assuming there is no concept shift(\textit{i.e.}, $p(y|x)=q(y|x)$), according to Bayes' Law, joint probability distributions $p(x,y)$ and $q(x,y)$ can be aligned by drawing the marginal feature distributions $p(x)$ and $q(x)$ closer. Early methods \cite{mmd1, mmd2} attempt to learn domain-invariant representations in a reproducing kernel Hilbert space (RKHS) by minimizing Maximum Mean Discrepancy (MMD) \cite{mmd} or Wasserstein distance \cite{wasserstein} of the marginal distributions. Recently, inspired by generative adversarial networks \cite{gan}, adversarial methods has been proposed \cite{adda, adv1, adv2, adv3}. These methods typically utilize a domain discriminator to encourage domain confusion via an adversarial minimax game. However, they could not trivially generalize to open-set problem.

\textbf{Open Set Domain Adaptation} (OSDA). OSDA is a realistic setting where the target domain contains private samples that should be recognized as unknown. \cite{osda} is one of the pioneering attempts to tackle the open set domain adaptation. Then, a group of works are proposed to separate known from unknown while aligning the known classes in an  adversarial way. Typically, OSBP \cite{osbp} utilizes a domain discriminator to enable the feature generator to learn discriminative representations for known/unknown separation. Recently, ROS \cite{ros} employs self-supervised learning methods to enhance the performance of the classifier and achieves great performance. PSDC \cite{psdc} utilize a prototype neural network and an auxiliary dummy classifier for classification. However, despite the remarkable achievements on the OSDA problem, the potential label shift problem is not considered in all of these works.

\begin{figure*}[t]
  \centering
   \includegraphics[width=7in, trim={15mm 0 0 0}, clip]{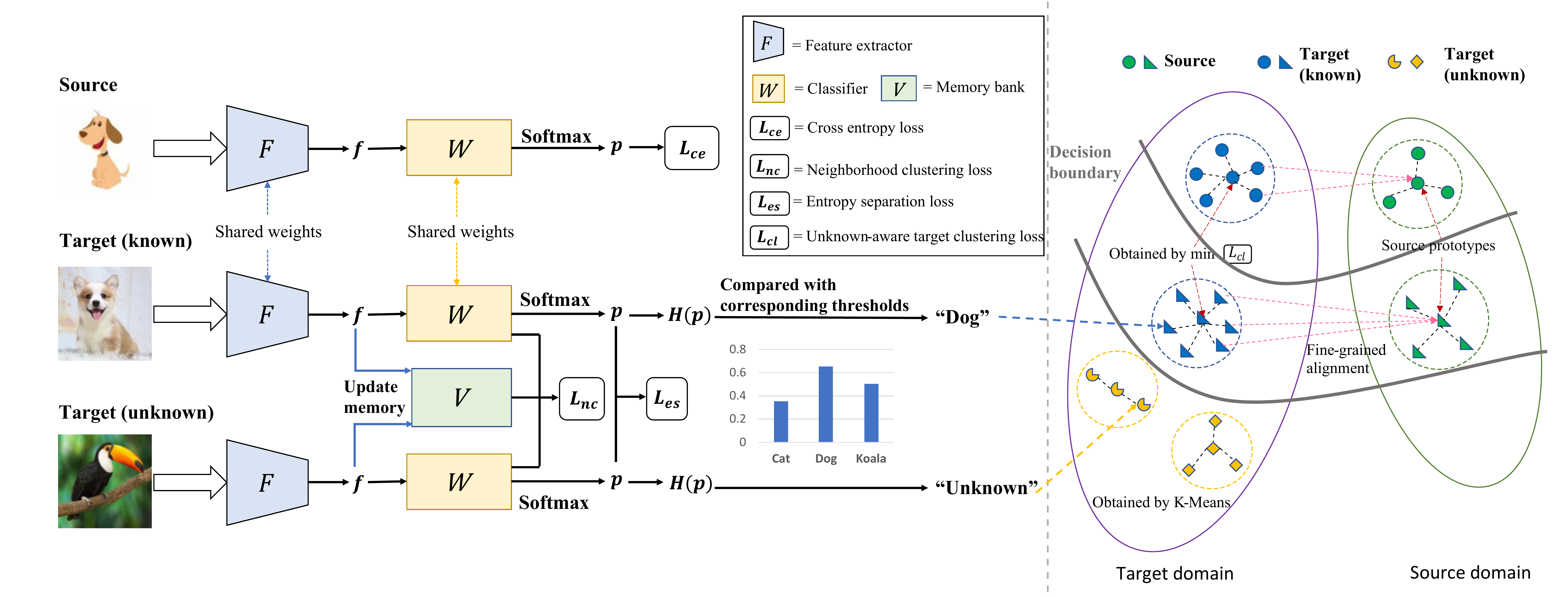}
   \caption{An illustration of our framework. Both source samples and target samples are passed through the feature extractor $F$ and the classifier $W$. The cross entropy loss is used to minimize the classification error on source samples to form source prototypes. For each mini-batch target domain samples, we update their features in the memory bank $V$, then we calculate their neighborhood clustering loss and entropy separation loss. After recognizing unknown samples using sample-wise thresholds, the target features are utilized to calculate unknown-aware target clustering loss to form tight clusters. Then we gradually align the two domains to mitigate the domain shift. }
   \label{method}

\end{figure*}

\textbf{Universal Domain Adaptation} (UniDA). UniDA \cite{unida} is a more challenging setting which requires no prior knowledge between the source and the target domain and allows both domains to have private samples. DANCE \cite{dance} utilizes self-supervision to encourage target samples to align with their neighbors or source prototypes. Ovanet \cite{ova} automatically generate thresholds without hyperparameters to effectively distinguish unknown samples. In fact, our setting could also be seen as an extension of UniDA, which makes IOSDA more important. 

\textbf{Imbalanced Domain Adaptation}. Label shift is a common problem in real life. As to alleviate the label shift, \cite{bbse} proposed a distribution estimator to detect it and correct the classifier by reweighting source samples. However, this method did not consider covariate shift. Then, \cite{yjf, coal} introduce label shift to common UDA problem by proposing Imbalanced Domain Adaptation (IDA) where $p(y) \neq q(y)$, $p(x|y) \neq q(x|y)$ and $p(y|x) = q(y|x)$. IDA could be seen as an extension of Partial Domain Adaptation (PDA) \cite{partial, pda1} where the source domain has private classes. In fact, the OSDA setting can also be seen as a special case of IDA, but IDA methods cannot trivially generalize to OSDA setting.

\textbf{Domain Adaptation Under Open Set Label Shift}. Very recently, Garg {\textit{et al.}} \cite{osls} proposed Domain Adaptation Under Open Set Label Shift (OSLS), which assumes that $p(y) \ne q(y)$ but $p(x|y) = q(x|y)$ and there are private samples in the target domain. They then translate the OSLS problem into a combination of Positive and Unlabeled (PU) learning problem and label shift problem to solve it. This setting is similar to ours, but we also assume the existence of covariate shift, which is more practical. Intuitively, OSLS method tends to recognize more samples as unknown when the marginal feature probability gap is large.

To sum up, there are five similar but different problem settings. Vanilla closed-set UDA assumes $p(y)=q(y),p(x|y) \neq q(x|y)$. Based on this, OSDA assumes the target domain has private classes. UniDA introduces open-partial DA (OPDA) where both domains have private classes. However, both settings assume $p(y)=q(y)$ for shared classes, which is relaxed by IDA. OSLS assumes the target domain has private categories and $p(y)\ne q(y)$, but it leaves the class-conditional distribution be the same, {\textit{i.e.}},$p(x|y) = q(x|y)$ for shared classes.

\section{Methodology}

\label{me}
\subsection{Problem Setting and Overall Idea}
Let $\mathcal{X}$ $\subseteq$ $\mathbb{R} ^ d$ be the feature space of the input images and $\mathcal{Y}=\{1,2,3,...,K,K+1\}$ be the output space of the classifier. Let $p$ and $q$ be the source and target distributions defined on $\mathcal{X} \times \mathcal{Y}$. In Imbalanced Open Set Domain Adaptation, we are given a source domain $\mathcal{D}^s=\{(x_i^s,y_i^s)\}_{i=1}^{N_s}\sim p$ and a target domain $\mathcal{D}^t=\{x_i^t\}_{i=1}^{N_t}\sim q$.
Denote by $\mathcal{C}_t=\mathcal{Y}$ and $\mathcal{C}_s=\mathcal{Y} - \{K+1\}$ the label sets of the target domain and the source domain, respectively. IOSDA assumes that the conditional feature distributions and the label distributions are different:
\begin{equation}
  \begin{cases}
    p(x|y=j) \ne q(x|y=j), \forall j \in \mathcal{C}_s \\
    p(y=j) \ne q(y=j), \forall j \in \mathcal{C}_s  \\
    p(y=K+1) = 0, q(y=K+1) \ne 0
  \end{cases}
\end{equation}

The goal of IOSDA is to train a model on $\mathcal{D}^s \cup \mathcal{D}^t$ to correctly classify samples of the target domain into one of the shared classes or the ``unknown'' class. In order to achieve this, we use DANCE \cite{dance} as the basic framework as it is a UniDA method and utilizes prototypical networks which could be helpful in class-imbalanced problems. Based on DANCE, moving-threshold estimation is used to effectively distinguish unknown samples while aligning the two domains in a gradual and fine-grained manner. Then we utilize unknown-aware target clustering loss to form tight clusters to alleviate the label shift problem. Our method is illustrated in \cref{method}.

\subsection{Learning Source Prototypes}
Inspired by \cite{minimax} and \cite{proto_uda}, our model contains a deep convolutional neural network as the feature extractor \textit{F} and a fully-connected layer as the classifier. The output of \textit{F} is $\ell_2$-normalized before sending to the classifier. The weight of the classifier is denoted by ${\boldsymbol{W}} = [{\boldsymbol{w}}_1, {\boldsymbol{w}}_2,...,{\boldsymbol{w}}_K] \in \mathbb{R} ^ {d \times K}$.

We first adopt a prototypical network \cite{proto1} that can bring source domain samples closer to their corresponding class centroids. Given source sample features $x_i^s \in \mathbb{R} ^ d$ and labels $y_i^s$, we get source prototypes using cross entropy loss:
\begin{equation}
  \label{l_ce}
  \mathcal{L}_{ce} = \mathbb{E}_{(x_i^s,y_i^s) \sim p} \sum_{k=1}^{K} -\ln {\boldsymbol{p}}_{ik}^{s} \cdot \mathbbm{1}_{\{y_i^s=k\}}
\end{equation}
where
\begin{equation}
  \label{temp}
  {\boldsymbol{p}}_{ik}^{s}= \frac{{\rm{exp}}({\boldsymbol{w}}_k^{\top}x_i^s/\tau)}{\sum_{j=1}^{K}{\rm{exp}}({\boldsymbol{w}}_j^{\top}x_i^s/\tau)}
\end{equation}
is the predicted probability for $x_i^s$ to be in class $k$ and the temperature parameter $\tau$ controls the concentration degree \cite{tao}. The columns of ${\boldsymbol{W}}$ can be seen as the source prototypes \cite{proto_uda}. In this way, we avoid additional parameters and computation when calculating source prototypes.

\subsection{Moving-threshold Estimation}
The key challenge of open set domain adaptation is how to separate common samples from private samples in the target domain. Assuming the model has more confidence (lower entropy) in shared-class samples than unknown samples, one mainstream method in OSDA is to set a confidence threshold on the entropy of the output of the classifier. Because the entropy of unknown samples is usually larger than known ones', we could set a pre-defined threshold $\rho$ and recognize those samples whose entropy is larger than $\rho$ as unknown. Following \cite{dance}, $\rho$ is set to be ${{\rm{ln}}(K)}/2$. However, the threshold leads the model to be sensitive to hyperparameter tuning that may influence the robustness of the model in real world scenarios. We tackle this problem by proposing the moving-threshold estimation method. 

One important drawback of the previous methods is that they ignore the semantic knowledge within unknown samples. In fact, there are various kinds of unknown images in an open world that could become unknown target samples. As a result, the relationships between unknown classes and known classes can vary greatly. Some unknown classes may be far from all known classes, and we call these classes ``easy unknown'' classes. Meanwhile, some unknown classes may be semantically close to certain known classes, and we call these classes ``hard unknown'' classes.  As a result, the model may make overconfident predictions for those hard unknown samples that are similar to one of the known classes. On the other hand, for some classes that containing more ``easy unknown'' samples, they may prefer a higher threshold to prevent known samples from being mistaken as unknown, and it won't hurt the performance because ``easy unknown'' samples have higher entropy. In other words, we need sample-wise thresholds rather than one general threshold for all classes. 

In order to automatically generate sample-wise thresholds, the pseudo-labels of target samples are used in a self-supervised way. Let $T_i$ be the set of indices of target samples that are recognized as class $i$. We do not distinguish unknown samples in this step, therefore $T_1 \cup T_2 \cup ... \cup T_K=\{1,2,...,N_t\}$. As ``easy unknown'' samples have higher entropy than ``hard unknown'' samples, we can infer which classes contain more ``easy unknown'' samples or more ``hard unknown'' samples by the mean entropy of each class. Higher entropy indicates more ``easy unknown'' samples and vice versa. Concretely, let ${\boldsymbol{p}}$ be the output of the classifier after softmax and ${\boldsymbol{p}_i}$ be the $i$-th target sample's probability distribution, we calculate class-wise entropy ${\boldsymbol{E}}$ by
\begin{equation}
  {\boldsymbol{E}}_i=\frac{1}{|T_i|}\sum_{j=1}^{|T_i|}H({\boldsymbol{p}}_{T_{ij}}), \forall i \in \mathcal{C}_s
\end{equation}
where $T_{ij}$ denotes the $j$-th value of $T_i$. Then we change thresholds according to the class-wise entropy. Let $r$ be the moving-threshold ratio, we calculate the thresholds ${\boldsymbol{q}}$ by
\begin{equation}
  \label{q}
  {\boldsymbol{q}}_i=(0.5 - r + 2 \cdot \frac{{\boldsymbol{E}}_i - \min({\boldsymbol{E}})}{\max({\boldsymbol{E}}) - \min({\boldsymbol{E}})} \cdot r) \ln K
\end{equation}
The thresholds are then used to distinguish unknown samples. For the $i$-th target sample, its predicted label $\hat{y}_{i}^{t}$ is
\begin{equation}
  \hat{y}_{i}^{t}=
  \begin{cases}
    K+1, & H({\boldsymbol{p}}_i) > {\boldsymbol{q}}_{\mathop{\arg\max}\limits_{j} {\boldsymbol{p}_{ij}}} \\
    \mathop{\arg\max}\limits_{j} {\boldsymbol{p}_{ij}}, & \text{otherwise}
  \end{cases}
\end{equation}
In this way, we could set a proper threshold for each target sample rather than using one for all samples. Moreover, in order to encourage the model to give clearer decisions, entropy separation loss \cite{dance} is utilized to refine the decision boundaries. Let $m$ be the confidence interval and $B_t$ be the corresponding indices of each target domain mini-batch, entropy separation loss is defined as:
\begin{equation}
  \mathcal{L}_{es}=\frac{1}{|B_t|}\sum_{i \in B_t}\mathcal{L}_{es}({\boldsymbol{p}}_i)
\end{equation}
where
\begin{equation}
  \label{l_es}
  \mathcal{L}_{es}({\boldsymbol{p}}_i)=
  \begin{cases}
    -|H({\boldsymbol{p}}_i)-\rho|, & |H({\boldsymbol{p}}_i)-\rho| \ge m \\
    0, & \text{otherwise}
  \end{cases}
\end{equation}
The confidence interval $m$ enables entropy separation loss to keep target samples away from the known/unknown decision boundary. 

\begin{figure*}[!t]
  \centering
  \subfloat[]{\includegraphics[width=3in]{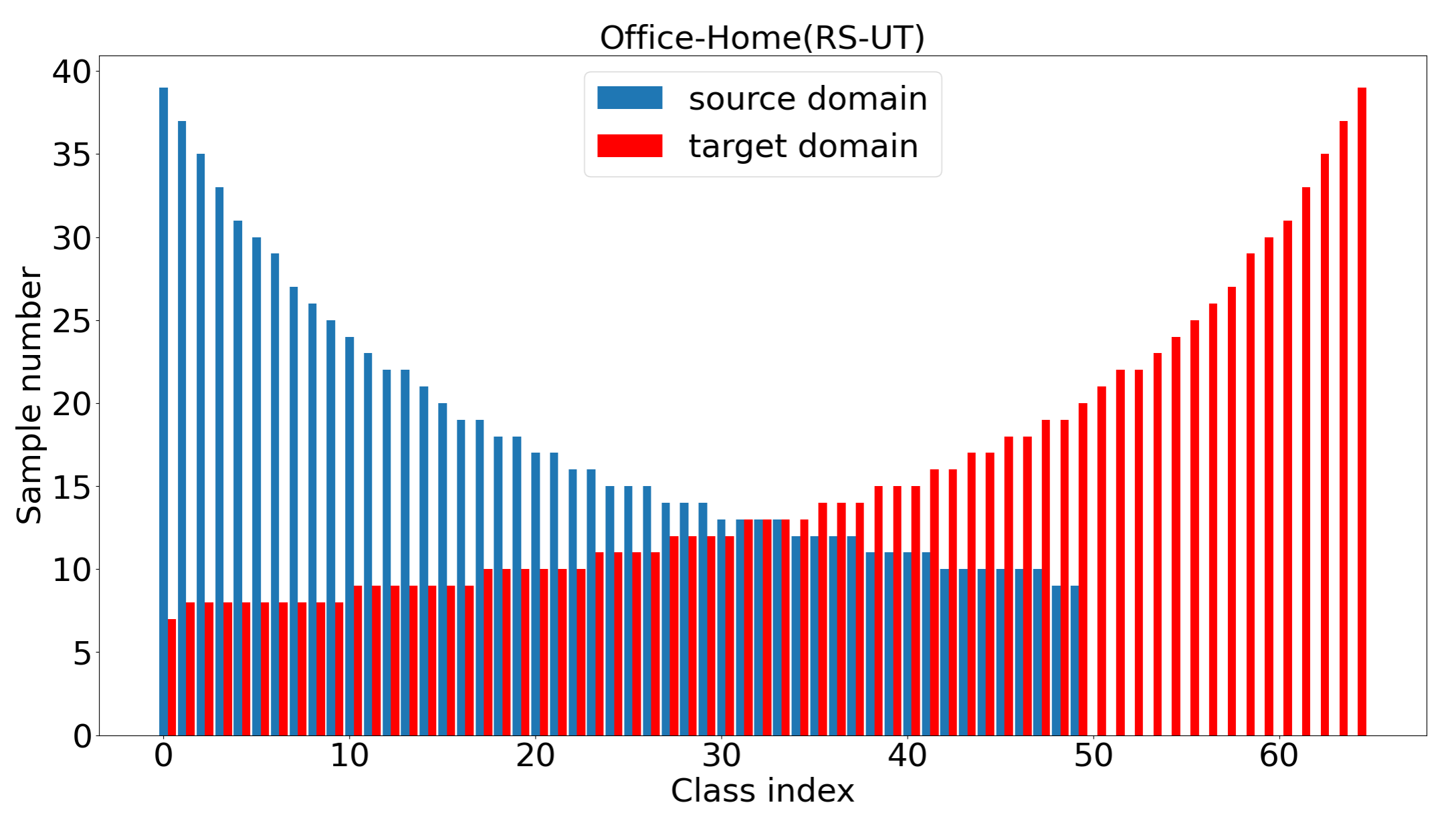}%
  \label{dataset.oh}}
  \hfil
  \subfloat[]{\includegraphics[width=3.2in]{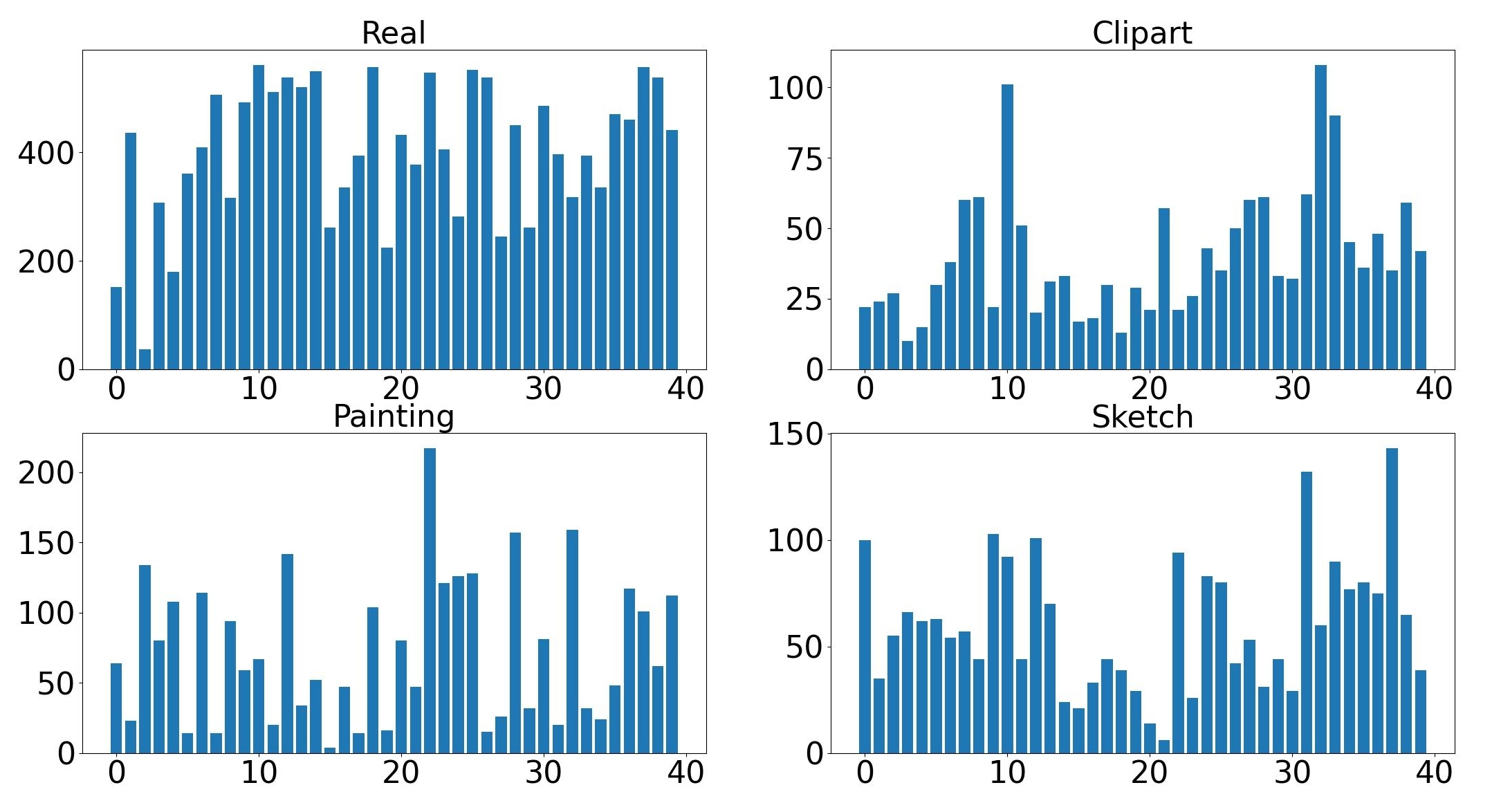}%
  \label{dataset.dn}}
  \caption{Label distributions of the adopted datasets. (a) The Office-Home dataset is sub-sampled with {\textbf{R}}eversely-unbalanced {\textbf{S}}ource and {\textbf{U}}nbalanced {\textbf{T}}arget (RS-UT) protocol. The label distributions of VisDA-C are similar to this protocol. (b) Label distributions of the DomainNet dataset.}
  \label{dataset}
\end{figure*}

\subsection{Unknown-aware Target Clustering Loss}
To effectively handle class imbalance, we apply clustering-based methods. By forming tighter clusters in the latent space of the target domain, the class imbalance problem can be effectively alleviated \cite{lmle, isfda}. Inspired by contrastive loss \cite{kcl}, we design the unknown-aware target clustering loss to minimize the distance between similar samples in one mini-batch and eventually form tighter clusters.

Given two samples $x_{i}^{t}$ and $x_{j}^{t}$ from a target domain mini-batch and their probability distributions ${\boldsymbol{p}_{i}}$ and ${\boldsymbol{p}_{j}}$, we wish to pull them closer if they have the same pseudo-label. For this purpose, KL-divergence is utilized to measure the distance between ${\boldsymbol{p}_{i}}$ and ${\boldsymbol{p}_{j}}$:
\begin{equation}
  \label{l_kl}
  \mathcal{L}_{KL}({\boldsymbol{p}_{i}}, {\boldsymbol{p}_{j}}) = \mathcal{D}_{KL}({\boldsymbol{p}_{i}} || {\boldsymbol{p}_{j}}) + \mathcal{D}_{KL}({\boldsymbol{p}_{j}} || {\boldsymbol{p}_{i}})
\end{equation}

However, the unknown samples may come from various semantic classes, and directly minimizing the distance between them may confuse the classifier. In other words, only samples that have similar semantic meanings should be aligned. Therefore, to pull unknown samples closer properly, we first store all $\ell_2$-normalized features of target samples via a memory bank ${\boldsymbol{V}}=[{\boldsymbol{V}}_1, {\boldsymbol{V}}_2,...,{\boldsymbol{V}}_{N_t}] \in \mathbb{R} ^ {d \times N_t}$. In every iteration, the up-to-date features are updated in the memory bank. Then we perform K-means clustering in ${\boldsymbol{V}}$ to form $Z$ clusters $S \in \mathbb{R} ^ {d \times Z}$. For each unknown sample $x_{t}$, its corresponding cluster index $z$ can be calculated as:
\begin{equation}
  z = \mathop{\arg\min}\limits_{i} {||x_{t} - S_{i}||}_{2}
\end{equation}
where $S_{i}$ is the $i$-th column of $S$. Different clusters indicate different semantic classes, and the unknown samples within the same cluster are pulled closer. Thus, the pseudo-labels of unknown samples should be $K+1+z, z \in \{0,1,...,Z-1\}$. Furthermore, since the pseudo-labels may be incorrect, directly applying the KL-loss may form unbefitting clusters. Therefore, different weights should be assigned to different pairs of samples. Pairs with higher confidence score samples should be assigned higher weights, and vice versa. For target samples recognized as known classes, the largest logit of their probability distributions is treated as the confidence score. For unknown samples, the entropy of their probability distributions is set to be the confidence score. Therefore, for each pair of samples $(x_{i}^{t}, x_{j}^{t})$ in a mini-batch, the corresponding weight can be calculated as:
\begin{equation}
  w_{ij}=
  \begin{cases}
    \min(c_i, c_j), & \hat{y}_{i}^{t} = \hat{y}_{j}^{t} \\
    0, & \text{otherwise}
  \end{cases}
\end{equation}
where $(c_i, c_j)$ are the confidence scores of $(x_{i}^{t}, x_{j}^{t})$. Note that we do not maximize the distance between dissimilar samples as this may lead to overconfident predictions under huge domain gaps. The final unknown-aware target clustering loss can be formulated as:
\begin{equation}
  \label{l_cl}
  \mathcal{L}_{cl}({\boldsymbol{p}_{i}}, {\boldsymbol{p}_{j}})=\frac{1}{|B_t|^{2}} \sum_{i \in B_t} \sum_{j \in B_t, j \ne i} w_{ij} \cdot \mathcal{L}_{KL}({\boldsymbol{p}_{i}}, {\boldsymbol{p}_{j}})
\end{equation}

By utilizing the unknown-aware target clustering loss, our model could learn tighter clusters in the target domain that could improve the robustness under class imbalance problem.

\begin{algorithm}
  \caption{Training Procedure of OMEGA}\label{algorithm}
  \KwIn{Source domain $(x_i^s, y_i^s)_{i=1}^{N_s}$, target domain $(x_i^t)_{i=1}^{N_t}$, prototypical network $f_{\theta}$, maximum epoch number $N_b$}
  \For{each epoch in range(0,$N_b$)} {
    Randomly retrieve mini-batch source domain samples $x^s$ and mini-batch target domain samples $x^t$\;
    Calculate the cross-entropy loss $\mathcal{L}_{ce}$ of $x^s$ using Eq. \ref{l_ce}\;
    Calculate the neighborhood clustering loss $\mathcal{L}_{nc}$ of $x^t$ using Eq. \ref{l_nc}\;
    Calculate the entropy separation loss $\mathcal{L}_{es}$ of $x^t$ using Eq. \ref{l_es}\;
    Calculate the unknown-aware target clustering loss $\mathcal{L}_{cl}$ of $x^t$ using Eq. \ref{l_cl}\;
    Backpropagate the total loss $\mathcal{L}$ using Eq. \ref{all}\;
    Update target domain clusters using K-means\;
    Update class-specific thresholds using Eq. \ref{q}\;
  }
  \KwOut{The trained model $f_{\theta}$}
\end{algorithm}

\subsection{Gradual Alignment}
Common OSDA methods usually improve the model accuracy by mitigating the distribution shift between the source and target domain in an adversarial way. However, in IOSDA, the domain gap is larger than usual, and forcefully aligning the two distributions may lead to catastrophic misalignment as the pseudo-labels are more likely to be incorrect. Therefore, following DANCE, we utilize the domain-specific batch normalization \cite{dsbb} to achieve weak alignment. Furthermore, neighborhood clustering loss \cite{dance} is adopted to gradually align the two domains. Denoting ${\boldsymbol{F}}=[{\boldsymbol{V}}_1,...,{\boldsymbol{V}}_{N_t}, {\boldsymbol{w}}_1,...,{\boldsymbol{w}}_K]$, neighborhood clustering loss is calculated as
\begin{equation}
  \label{l_nc}
  \mathcal{L}_{nc}=-\frac{1}{|B_t|}\sum_{i \in B_t}\sum_{j=1,j \ne i}^{N_t+K}p_{ij}{\rm{ln}}(p_{ij})
\end{equation}
where
\begin{equation}
  p_{ij}=\frac{{\rm{exp}}({\boldsymbol{F}}_j^{\top}x_i^t/\tau)}{\sum_{l=1,l \ne i}^{N_t+K}{\rm{exp}}({\boldsymbol{F}}_l^{\top}x_i^t/\tau)}
\end{equation}

\begin{table*}
  \begin{centering}
    \caption{Results (\%) on \textbf{Office-Home (RS-UT)} (ResNet-50).}
    \label{result_oh}
    \scalebox{0.75}{
    \begin{tabular}{|l|ccc|ccc|ccc|ccc|ccc|lll|lll|}
      \hline
      \multirow{2}{*}{Methods}            & \multicolumn{3}{c|}{Rw$\rightarrow$Pr}                                              & \multicolumn{3}{c|}{Rw$\rightarrow$Cl}                                              & \multicolumn{3}{c|}{Pr$\rightarrow$Rw}                                              & \multicolumn{3}{c|}{Pr$\rightarrow$Cl}                                              & \multicolumn{3}{c}{Cl$\rightarrow$Rw}                                                & \multicolumn{3}{|c|}{Cl$\rightarrow$Pr}    & \multicolumn{3}{c|}{Avg}       \\       
                                          & \multicolumn{1}{l}{OS*} & \multicolumn{1}{l}{UNK} & \multicolumn{1}{l|}{HOS}        & \multicolumn{1}{l}{OS*} & \multicolumn{1}{l}{UNK} & \multicolumn{1}{l|}{HOS}        & \multicolumn{1}{l}{OS*} & \multicolumn{1}{l}{UNK} & \multicolumn{1}{l|}{HOS}        & \multicolumn{1}{l}{OS*} & \multicolumn{1}{l}{UNK} & \multicolumn{1}{l|}{HOS}        & \multicolumn{1}{l}{OS*} & \multicolumn{1}{l}{UNK} & \multicolumn{1}{l|}{HOS}    & \multicolumn{1}{l}{OS*} & \multicolumn{1}{l}{UNK} & \multicolumn{1}{l|}{HOS}            & \multicolumn{1}{l}{OS*} & \multicolumn{1}{l}{UNK} & \multicolumn{1}{l|}{HOS}            \\ \hline
      STA \cite{sta}     & 63.43                   & 58.32                   & 60.77                           & 35.81                   & 72.41                   & 47.92                           & 64.78                   & 65.39                   & 65.08                           & 37.07                   & 72.41                   & 49.04                           & 50.83                   & 58.70                   & 54.48                                              &49.45 & 62.26 & 55.12          & 50.22 & 64.92 & 55.40          \\         
      OSBP \cite{osbp}   & 66.45                   & 54.89                   & 60.12                           & 41.89                   & 58.49                   & 48.82                           & 66.19                   & 60.80                   & 63.38                           & 38.03                   & 59.67                   & 46.45                           & 59.47                   & 53.73                   & 56.45                                             &55.56 & 60.48 & 57.91          & 54.60 & 58.01 & 55.52          \\
      ROS \cite{ros}     & 66.60                   & 39.52                   & 49.61                           & 43.53                   & 44.10                   & 43.81                           & 63.89                   & 47.23                   & 54.43                           & 41.53                   & 45.05                   & 43.18                           & 58.57                   & 34.42                   & 43.49                                              &51.85 & 26.42 & 35.13          & 54.33 & 39.46 & 44.94          \\
      UAN \cite{unida}   & 9.19                    & 33.67                   & 14.45                           & 3.23                    & 51.88                   & 6.08                            & 7.22                    & 17.20                   & 10.18                           & 2.28                    & 8.49                    & 3.60                            & 4.81                    & 9.37                    & 6.36                                               &6.09  & 13.72 & 8.44           & 5.47  & 22.39 & 8.19           \\
      DANCE \cite{dance} & 66.32                   & 57.18                   & 61.41                           & 42.40                   & 67.63                   & 52.12                           & 64.07                   & 64.43                   & 64.25                           & 46.70                   & 66.27                   & 54.79                           & 57.09                   & 62.14                   & 59.51                                              &51.07 & 68.50 & 58.51          & 54.61 & 64.36 & 58.43          \\
      CMU \cite{cmu}     & 50.50                   & 56.42                   & 53.30                           & 31.77                   & 54.48                   & 40.14                           & 45.64                   & 59.08                   & 51.50                           & 24.94                   & 61.79                   & 35.54                           & 30.64                   & 61.95                   & 41.00                                              &25.87 & 62.39 & 36.57          & 34.89 & 59.35 & 43.01          \\
      Ovanet \cite{ova}  & 67.34                   & 60.23                   & 63.58                           & 35.66                   & 77.59                   & 48.87                           & 63.51                   & 74.00                   & \textbf{68.36} & 32.70                   & 83.73                   & 47.03                           & 51.25                   & 78.39                   & 61.98                                                               &46.86 & 72.30 & 56.86          & 49.55 & 74.37 & 57.78          \\
      PULSE \cite{osls}  & 0.0                     & 100.0                   & 0.0                             & 0.0                     & 100.0                   & 0.0                             & 13.15                   & 100.0                   & 23.25                           & 0.0                     & 100.0                   & 0.0                             & 0.0                     & 100.0                   & 0.0                                         &1.25  & 99.87 & 2.48           & 2.40  & 99.98 & 5.15           \\ \hline
      \textbf{OMEGA}     & 59.61                   & 77.89                   & \textbf{67.53} & 42.23                   & 69.58                   & \textbf{52.56} & 62.75                   & 73.80                   & 67.83                           & 45.81                   & 69.81                   & \textbf{55.32} & 57.35                   & 68.83                   & \textbf{62.57}                                                                                 &58.49 & 64.04 & \textbf{61.14} & 54.37 & 70.66 & \textbf{61.16} \\ \hline
      \end{tabular}
    }
  \end{centering}
\end{table*}

\begin{table*}[]
  \begin{center}
  \caption{Results (\%) on \textbf{VisDA-C (RS-UT)} (ResNet-50).}
  \label{result_visda}
  \scalebox{0.75}{
  \begin{tabular}{|l|ccc|ccc|ccc|ccc|ccc|ccc|ccc|}
  \hline
  \multirow{2}{*}{Method} & \multicolumn{3}{c|}{$\omega=100, K=9$}                                              & \multicolumn{3}{c|}{$\omega=50, K=9$}                                               & \multicolumn{3}{c|}{$\omega=10, K=9$}                                               & \multicolumn{3}{c|}{$\omega=100, K=10$}                                             & \multicolumn{3}{c}{$\omega=50, K=10$}           & \multicolumn{3}{|c|}{$\omega=10, K=10$}                                              & \multicolumn{3}{c|}{Avg}                                            \\
                          & \multicolumn{1}{l}{OS*} & \multicolumn{1}{l}{UNK} & \multicolumn{1}{l|}{HOS} & \multicolumn{1}{l}{OS*} & \multicolumn{1}{l}{UNK} & \multicolumn{1}{l|}{HOS} & \multicolumn{1}{l}{OS*} & \multicolumn{1}{l}{UNK} & \multicolumn{1}{l|}{HOS} & \multicolumn{1}{l}{OS*} & \multicolumn{1}{l}{UNK} & \multicolumn{1}{l|}{HOS} & \multicolumn{1}{l}{OS*} & \multicolumn{1}{l}{UNK} & \multicolumn{1}{l}{HOS} &\multicolumn{1}{|l}{OS*} & \multicolumn{1}{l}{UNK} & \multicolumn{1}{l|}{HOS} & \multicolumn{1}{l}{OS*} & \multicolumn{1}{l}{UNK} & \multicolumn{1}{l|}{HOS} \\ \hline
  STA \cite{sta}                    & 37.00                   & 51.77                   & 43.16                    & 40.14                   & 47.64                   & 43.57                    & 45.97                   & 44.72                   & 45.33                    & 32.94                   & 59.90                   & 42.50                    & 36.02                   & 44.16                   & 39.67                   & 40.21                    & 22.57                   & 28.91                    & 38.71                   & 45.13                   & 40.52  \\
  OSBP \cite{osbp}                   & 45.00                   & 40.76                   & 42.77                    & 52.12                   & 37.68                   & 43.74                    & 57.04                   & 36.33                   & 44.39                    & 45.28                   & 44.79                   & 45.03                    & 48.90                   & 46.99                   & 47.93                   & 51.18                    & 75.98                   & \textbf{61.16}           & 49.92                   & 47.09                   & 47.50 \\
  ROS \cite{ros}                    & 51.63                   & 39.85                   & 44.97                    & 49.17                   & 41.22                   & 44.85                    & 53.94                   & 45.82                   & 49.53                    & 42.96                   & 48.03                   & 45.35                    & 39.23                   & 43.01                   & 41.03                   & 66.12                    & 40.51                   & 50.23                    & 50.51                   & 43.07                   & 45.99 \\
  UAN \cite{unida}                    & 55.22                   & 46.09                   & 50.24                    & 56.98                   & 46.00                   & 50.90                    & 57.95                   & 41.38                   & 48.28                    & 46.22                   & 40.16                   & 42.98                    & 46.56                   & 41.94                   & 44.13                   & 47.91                    & 30.65                   & 37.38                    & 51.81                   & 41.04                   & 45.65 \\
  DANCE \cite{dance}                  & 45.30                   & 55.73                   & 49.98                    & 38.54                   & 51.35                   & 44.03                    & 40.00                   & 62.03                   & 48.64                    & 41.28                   & 42.23                   & 41.75                    & 35.96                   & 44.31                   & 39.70                   & 38.07                    & 45.91                   & 41.63                    & 39.86                   & 50.26                   & 44.29 \\
  CMU \cite{cmu}                    & 25.82                   & 21.75                   & 23.61                    & 28.57                   & 22.63                   & 25.26                    & 25.97                   & 32.65                   & 28.93                    & 21.56                   & 48.00                   & 29.75                    & 24.56                   & 36.67                   & 29.42                   & 22.58                    & 43.16                   & 29.65                    & 24.84                   & 34.14                   & 27.77 \\
  Ovanet \cite{ova}                 & 36.25                   & 64.38                   & 46.38                    & 37.05                   & 70.99                   & 48.69                    & 40.67                   & 68.45                   & 51.02                    & 36.53                   & 66.33                   & 47.11                    & 36.52                   & 76.26                   & 49.38                   & 33.33                    & 74.69                   & 46.10                    & 36.73                   & 70.18                   & 48.11 \\
  PULSE \cite{osls}                  & 0.0                     & 100.0                   & 0.0                      & 0.0                     & 100.0                   & 0.0                      & 0.0                     & 100.0                   & 0.0                      & 0.0                     & 100.0                   & 0.0                      & 0.0                     & 100.0                   & 0.0                     & 0.0                      & 100.0                   & 0.0                      & 0.0                     & 100.0                   & 0.0 \\ \hline
  \textbf{OMEGA}          & 39.74                   & 84.57                   & \textbf{54.07}           & 42.98                   & 76.44                   & \textbf{55.02}           & 45.18                   & 67.11                  & \textbf{54.00}           & 37.62                   & 66.20                   & \textbf{47.97}           & 42.83                   & 66.63                   & \textbf{52.14}          & 41.82                    & 59.40                   & 49.08                    & 41.69                   & 70.05                   & \textbf{52.05} \\ \hline
  \end{tabular}
  }
  \end{center}
\end{table*}

By minimizing the entropy of the similarity between each mini-batch sample and all the other target samples and source prototypes, neighborhood clustering loss aligns each target sample to a source prototype or its neighbor in the target domain.

The overall objective can be formulated as:
\begin{equation}
  \label{all}
  \mathcal{L} = \mathcal{L}_{ce} + \eta_{1} (\mathcal{L}_{nc} + \mathcal{L}_{es}) + \eta_{2} \mathcal{L}_{cl}
\end{equation} 
where $\eta_{1}$ and $\eta_{2}$ are the trade-off parameters.

\section{Experiments}
\label{ex}

\begin{table*}[]
  \begin{center}
  \caption{Results (\%) on \textbf{DomainNet} (ResNet-50).}
  \label{result_dn}
  \scalebox{0.75}{
  \begin{tabular}{|l|ccc|ccc|ccc|ccc|ccc|ccc|}
  \hline
  \multirow{2}{*}{Methods} & \multicolumn{3}{c|}{R$\rightarrow$C}       & \multicolumn{3}{c|}{R$\rightarrow$S}       & \multicolumn{3}{c|}{R$\rightarrow$P}       & \multicolumn{3}{c|}{C$\rightarrow$R}       & \multicolumn{3}{c|}{C$\rightarrow$S}       & \multicolumn{3}{c|}{C$\rightarrow$P}       \\
                           & OS*   & UNK   & HOS            & OS*   & UNK   & HOS            & OS*   & UNK   & HOS            & OS*   & UNK   & HOS            & OS*   & UNK   & HOS            & OS*   & UNK   & HOS            \\ \hline
  STA \cite{sta}                     & 54.45 & 60.89 & 57.49          & 55.81 & 59.30 & 57.51          & 68.19 & 55.13 & 60.97          & 79.10 & 51.25 & 62.20          & 53.40 & 58.91 & 56.02          & 55.70 & 58.94 & 57.28          \\
  OSBP \cite{osbp}                    & 55.25 & 68.77 & 61.27          & 54.12 & 69.75 & 60.95          & 65.69 & 68.91 & 67.26          & 80.51 & 66.11 & 72.61          & 58.47 & 49.88 & 53.84          & 56.95 & 52.08 & 54.41          \\
  ROS \cite{ros}                     & 63.66 & 47.09 & 54.14          & 54.59 & 49.56 & 51.95          & 64.89 & 50.14 & 56.57          & 64.12 & 80.08 & 71.22          & 46.33 & 72.28 & 56.47          & 51.82 & 69.44 & 59.35          \\
  UAN \cite{unida}                     & 48.17 & 42.63 & 45.23          & 43.26 & 61.59 & 50.83          & 50.39 & 54.25 & 52.25          & 67.12 & 12.07 & 20.47          & 37.59 & 28.74 & 32.57          & 39.93 & 34.07 & 36.77          \\
  DANCE \cite{dance}                   & 57.25 & 63.69 & 60.30          & 54.97 & 65.64 & 59.83          & 59.43 & 74.19 & 66.00          & 66.81 & 75.40 & 70.84          & 47.68 & 76.01 & 58.60          & 60.68 & 60.06 & 60.37          \\
  CMU \cite{cmu}                     & 43.01 & 57.47 & 49.20          & 53.44 & 44.02 & 48.27          & 52.53 & 52.79 & 52.66          & 48.34 & 55.08 & 51.49          & 41.86 & 44.42 & 43.10          & 33.57 & 51.26 & 40.57           \\
  Ovanet \cite{ova}                  & 55.65 & 75.41 & 64.04          & 56.50 & 72.37 & 63.46          & 67.04 & 68.62 & 67.82          & 61.76 & 83.75 & 71.09          & 55.25 & 69.20 & 61.44          & 60.68 & 71.67 & \textbf{65.72} \\
  PULSE \cite{osls}                   & 0.0   & 100.0 & 0.0            & 0.0   & 100.0 & 0.0            & 0.0   & 100.0 & 0.0            & 0.31  & 99.89 & 0.61           & 0.0   & 100.0 & 0.0            & 0.0   & 100.0 & 0.0            \\ \hline
  \textbf{OMEGA}                   & 61.74 & 77.07 & \textbf{68.56} & 61.35 & 69.36 & \textbf{65.11} & 69.42 & 70.32 & \textbf{69.87} & 75.93 & 79.92 & \textbf{77.88} & 54.33 & 74.74 & \textbf{62.92} & 60.40 & 71.44 & 65.46          \\ \hline
  \end{tabular}
  }
  \end{center}
  \vspace{-5mm}
\end{table*}

\begin{table*}[]
  \begin{center}
    \scalebox{0.75}{
      \begin{tabular}{|ccc|ccc|ccc|ccc|ccc|ccc|ccc|}
        \hline
        \multicolumn{3}{|c|}{S$\rightarrow$R}       & \multicolumn{3}{c|}{S$\rightarrow$C}       & \multicolumn{3}{c|}{S$\rightarrow$P}       & \multicolumn{3}{c|}{P$\rightarrow$C}       & \multicolumn{3}{c|}{P$\rightarrow$S}       & \multicolumn{3}{c|}{P$\rightarrow$R}       & \multicolumn{3}{c|}{Avg} \\
        OS*   & UNK   & HOS            & OS*   & UNK   & HOS            & OS*   & UNK   & HOS            & OS*   & UNK   & HOS            & OS*   & UNK   & HOS            & OS*   & UNK   & HOS            & OS*    & UNK    & HOS   \\ \hline
        84.42 & 58.42 & 69.06          & 62.42 & 56.02 & 59.05          & 65.15 & 60.59 & 62.79          & 52.86 & 51.76 & 52.31          & 55.45 & 48.93 & 51.99          & 77.82 & 38.65 & 51.64          & 63.73  & 54.90  & 58.19 \\
        82.26 & 65.73 & 73.07          & 62.48 & 48.34 & 54.51          & 67.08 & 67.50 & 67.29          & 53.95 & 52.48 & 53.21          & 59.15 & 59.61 & 59.38          & 78.01 & 59.05 & 67.22          & 64.49  & 60.68  & 62.09 \\
        70.14 & 70.85 & 70.49          & 60.38 & 46.16 & 52.32          & 65.18 & 57.00 & 60.82          & 51.70 & 40.87 & 45.65          & 45.36 & 58.51 & 51.10          & 56.59 & 59.49 & 58.00          & 57.90  & 58.45  & 57.34 \\
        57.88 & 16.87 & 26.12          & 49.34 & 11.82 & 19.07          & 43.02 & 22.11 & 29.21          & 39.73 & 16.49 & 23.31          & 49.28 & 9.34  & 15.71          & 68.30 & 5.09  & 9.48           & 49.50  & 26.26  & 30.09 \\
        75.43 & 72.28 & 73.82          & 56.15 & 70.33 & 62.44          & 64.69 & 72.20 & 68.24          & 50.00 & 64.63 & 56.38          & 56.12 & 62.71 & 59.23          & 65.64 & 63.52 & 64.36          & 59.57  & 68.39  & 63.37 \\
        63.22 & 44.43 & 52.18          & 47.15 & 48.55 & 47.84          & 46.01 & 51.09 & 48.42          & 41.37 & 46.99 & 44.00          & 41.57 & 53.52 & 46.80          & 53.35 & 44.54 & 48.55          & 47.12  & 49.51  & 47.76 \\
        87.46 & 58.78 & 70.31          & 62.79 & 71.16 & 66.72          & 70.52 & 74.25 & 72.34          & 59.78 & 64.94 & 62.25          & 64.81 & 63.18 & 63.99          & 58.29 & 82.58 & 68.34          & 63.38  & 71.33  & 66.46 \\
        0.0   & 100.0 & 0.0            & 0.0   & 100.0 & 0.0            & 13.15 & 100.0 & 23.25          & 0.0   & 100.0 & 0.0            & 0.09  & 100.0 & 0.18           & 0.03  & 100.0 & 0.06           & 0.04   & 99.99  & 0.07  \\ \hline
        86.16 & 80.96 & \textbf{83.48} & 65.52 & 71.68 & \textbf{68.46} & 69.87 & 76.48 & \textbf{73.03} & 62.12 & 64.42 & \textbf{63.25} & 60.79 & 70.31 & \textbf{65.20} & 76.96 & 77.73 & \textbf{77.35} & 67.05  & 73.70  & \textbf{70.05} \\ \hline
        \end{tabular}
    }
  \end{center}
\end{table*}

\begin{table}[]
  \begin{center}
  \caption{HOS (\%) for ablation study on \textbf{VisDA-C (RS-UT)}.}
  \label{ablation}
  \scalebox{0.9}{
    \begin{tabular}{|l|c|c|c|}
      \hline
      Methods      & $\omega=100, K=9$   & $\omega=50, K=9$           & $\omega=10, K=9$            \\ \hline
      Baseline        & 49.98          & 44.03          & 48.64                    \\
      OMEGA w/o $\mathcal{L}_{cl}$ & 51.13          & 50.26          & 50.18                    \\
      OMEGA w/o ME & 52.78          & 53.83          & 52.77                    \\ \hline
      {\textbf{OMEGA}}        & \textbf{54.07} & \textbf{55.02} & \textbf{54.00}  \\ \hline
      \end{tabular}
  }
  \end{center}
  \vspace{-7mm}
\end{table}

\begin{table}[]
  \begin{center}
  \scalebox{0.9}{
    \begin{tabular}{|c|c|c|c|}
      \hline
      $\omega=100, K=10$         & $\omega=50, K=10$          & $\omega=10, K=10$          & Avg            \\ \hline
      41.75          & 39.70          & 41.63          & 44.29          \\
      44.48          & 44.68          & 43.33          & 47.34          \\
      46.58          & 49.71          & 45.75          & 50.24          \\ \hline
      \textbf{47.97} & \textbf{52.14} & \textbf{49.08} & \textbf{52.05} \\ \hline
      \end{tabular}
  }
  \end{center}
  \vspace{-4mm}
\end{table}

\begin{table*}[]
  \begin{center}
  \caption{HOS (\%) on \textbf{Office-Home} in \textbf{OSDA} setting. Following previous settings\cite{psdc,ros}, the first 25 classes in alphabetical order are treated as known classes. $^{*}$Cited from (Liu {\textit{et al.}} \cite{psdc}).}
  \label{result_osda}
  \scalebox{0.95}{
    \begin{tabular}{|l|l|l|l|l|l|l|l|l|l|l|l|l|l|}
      \hline
      Method & Pr$\rightarrow$Rw         & Pr$\rightarrow$Cl         & Pr$\rightarrow$Ar         & Ar$\rightarrow$Pr         & Ar$\rightarrow$Rw         & Ar$\rightarrow$Cl         & Rw$\rightarrow$Ar         & Rw$\rightarrow$Pr         & Rw$\rightarrow$Cl         & Cl$\rightarrow$Rw         & Cl$\rightarrow$Ar         & Cl$\rightarrow$Pr         & Avg           \\ \hline
      STA$^{*}$\cite{sta}    & 69.7          & 55.0          & 63.1          & 63.7          & 62.1          & 56.3          & 65.0          & 66.4          & 54.2          & 66.3          & 57.9          & 62.5          & 61.9          \\
      OSBP$^{*}$\cite{osbp}   & 73.9          & 53.2          & 63.2          & 65.2          & 72.9          & 55.1          & 66.7          & 72.3          & 54.5          & 70.6          & \textbf{64.3} & 64.7          & 64.7          \\
      ROS$^{*}$\cite{ros}    & \textbf{74.4} & 56.3          & 60.6          & \textbf{69.3} & 76.5          & 60.1          & \textbf{68.8} & \textbf{75.7} & \textbf{60.4} & 68.6          & 58.9          & 65.2          & 66.2          \\
      UAN$^{*}$\cite{unida}    & 0.2           & 0.0           & 0.0           & 0.0           & 0.2           & 0.0           & 0.2           & 0.1           & 0.0           & 0.2           & 0.0           & 0.2           & 0.1           \\
      DANCE\cite{dance}  & 68.4          & 51.2          & 62.8          & 64.3          & 70.1          & 58.2          & 61.2          & 71.9          & 54.8          & 66.6          & 59.6          & 62.7          & 62.6          \\
      CMU\cite{cmu}    & 54.6          & 40.0          & 41.2          & 51.9          & 56.0          & 45.4          & 53.5          & 56.8          & 45.6          & 52.1          & 41.4          & 47.2          & 48.8          \\
      Ovanet\cite{ova} & 69.7          & 52.6          & 60.1          & 66.2          & 69.2          & 58.5          & 67.8          & 67.2          & 58.7          & 68.9          & 60.8          & 64.3          & 63.7          \\
      PULSE\cite{osls}  & 27.3          & 15.7          & 43.1          & 39.8          & 41.4          & 25.0          & 33.7          & 36.8          & 27.5          & 29.8          & 39.0          & 31.8          & 32.6          \\ \hline
      OMEGA  & 72.8          & \textbf{57.6} & \textbf{65.5} & 68.9          & \textbf{76.6} & \textbf{60.9} & 67.7          & 73.5          & 59.4          & \textbf{70.8} & 62.2          & \textbf{65.6} & \textbf{66.8} \\ \hline
      \end{tabular}
  }
  \end{center}
\end{table*}

\begin{table*}[]
  \begin{center}
  \caption{HOS (\%) on \textbf{Office-Home} in \textbf{OPDA} setting. Following previous settings\cite{ova,dance}, the first 10 classes in alphabetical order are treated as shared classes, the following 5 classes are source private classes, and the rest of the classes are  target private classes. $^{*}$Cited from (Saito {\textit{et al.}} \cite{ova}).}
  \label{result_opda}
  \scalebox{0.95}{
    \begin{tabular}{|l|l|l|l|l|l|l|l|l|l|l|l|l|l|}
      \hline
      Method & Ar$\rightarrow$Cl           & Ar$\rightarrow$Pr           & Ar$\rightarrow$Rw           & Cl$\rightarrow$Ar           & Cl$\rightarrow$Pr           & Cl$\rightarrow$Rw           & Pr$\rightarrow$Ar           & Pr$\rightarrow$Cl           & Pr$\rightarrow$Rw           & Rw$\rightarrow$Ar           & Rw$\rightarrow$Cl           & Rw$\rightarrow$Pr           & Avg           \\ \hline
      STA\cite{sta}    & 39.1          & 43.6          & 47.5          & 42.7          & 44.0          & 43.7          & 46.3          & 41.4          & 47.0          & 49.7          & 45.2          & 53.5          & 45.3          \\
      OSBP\cite{osbp}   & 39.6          & 45.1          & 46.2          & 45.7          & 45.2          & 46.8          & 45.3          & 40.5          & 45.8          & 45.1          & 41.6          & 46.9          & 44.5          \\
      ROS$^{*}$\cite{ros}    & 41.2          & 44.7          & 48.3          & 45.9          & 47.1          & 48.3          & 44.8          & 42.1          & 49.6          & 46.4          & 39.9          & 48.2          & 45.5          \\
      UAN$^{*}$\cite{unida}    & 51.6          & 51.7          & 54.3          & 61.7          & 57.6          & 61.9          & 50.4          & 47.6          & 61.5          & 62.9          & 52.6          & 65.2          & 56.6          \\
      DANCE\cite{dance}  & 62.2          & 70.6          & 79.4          & 69.3          & 67.2          & 77.8          & 69.4          & 59.2          & 78.1          & 76.5          & 59.6          & 74.0          & 70.3          \\
      CMU$^{*}$\cite{cmu}    & 56.0          & 56.9          & 59.1          & 66.9          & 64.2          & 67.8          & 54.7          & 51.0          & 66.3          & 68.2          & 57.8          & 69.7          & 61.6          \\
      Ovanet$^{*}$\cite{ova} & 62.8          & 75.6          & 78.6          & 70.7          & 68.8          & 75.0          & 71.3          & 58.6          & 80.5          & 76.1          & 64.1          & 78.9          & 71.8          \\
      PULSE\cite{osls}  & 23.8          & 38.4          & 40.6          & 39.5          & 41.6          & 47.4          & 29.2          & 34.1          & 44.8          & 36.2          & 29.4          & 41.0          & 37.2          \\ \hline
      OMEGA  & \textbf{64.9} & \textbf{79.2} & \textbf{87.1} & \textbf{75.2} & \textbf{72.9} & \textbf{81.3} & \textbf{75.5} & \textbf{63.3} & \textbf{83.7} & \textbf{77.2} & \textbf{64.6} & \textbf{80.9} & \textbf{75.5} \\ \hline
      \end{tabular}
  }
  \end{center}
\end{table*}

\begin{table*}
  \caption{HOS (\%) in IOSDA/OPDA setting on Pr$\rightarrow$Rw task on Office-Home.}
  \label{zsens}
  \begin{center}
  \scalebox{1.30}{
     \begin{tabular}{l|l|l|l|l|l|l|l|}
        \cline{2-8}
                                           & $Z=0.2K$ & $Z=0.3K$ & $Z=0.4K$ & $Z=0.5K$      & $Z=0.6K$ & $Z=0.7K$ & $Z=0.8K$ \\ \hline
        \multicolumn{1}{|l|}{$\eta_2=0.1$} & 68.1/83.1     & 67.7/82.9     & 67.5/82.7     & \textbf{67.8/83.7} & 67.4/83.1     & 67.9/82.9     & 67.8/82.8     \\ \hline
        \multicolumn{1}{|l|}{$\eta_2=0.2$} & 70.6/83.5     & 69.4/84.1     & 69.2/84.2     & 69.3/84.5          & 69.3/84.6     & 69.3/84.0     & 69.4/84.7     \\ \hline
        \multicolumn{1}{|l|}{$\eta_2=0.3$} & 68.9/84.3     & 70.1/84.9     & 69.4/85.3     & 69.6/85.7          & 69.6/87.3     & 69.8/85.9     & 69.8/85.8     \\ \hline
        \multicolumn{1}{|l|}{$\eta_2=0.4$} & 68.1/83.9     & 69.1/84.2     & 70.3/84.7     & 70.6/85.3          & 69.8/85.6     & 69.6/85.5     & 69.4/86.0     \\ \hline
        \multicolumn{1}{|l|}{$\eta_2=0.5$} & 68.0/82.7     & 68.8/83.7     & 70.1/84.5     & 68.4/85.1          & 69.0/85.8     & 69.3/84.6     & 69.9/86.6     \\ \hline
     \end{tabular}
  }
  \vspace{-3mm}
  \end{center}
\end{table*} 

\subsection{Datasets}
We first construct the IOSDA datasets based on the imbalanced domain adaptation benchmarks created by Tan {\textit{et al.}} \cite{coal} and Li {\textit{et al.}} \cite{isfda} where Office-Home and VisDA follow RS-UT protocol and DomainNet exists salient heterogeneous label shift. The domain shift protocols are shown in \cref{dataset}.

{\textbf{Office-Home (RS-UT)}} is resampled based on Office-Home \cite{oh} where the source domain and the target domain are subject to two reverse Pareto distributions \cite{pareto}. Since the Art ({\textbf{Ar}}) domain is too small, Office-Home (RS-UT) only utilizes the other three domains: Clipart ({\textbf{Cl}}), Product (\textbf{Pr}) and Real-World(\textbf{Rw}). The last 15 classes in alphabetical order are treated as unknown classes to maintain enough images in the source domain. We take these three domains in turn as the source domain and target domain to build 6 tasks.

{\textbf{VisDA-C (RS-UT)}} is an imbalanced version of VisDA-C \cite{visda} created by Li {\textit{et al.}} following the same protocol as Office-Home (RS-UT). The challenging dataset contains three different imbalance factor $\omega=\frac{N_{max}}{N_{min}}$ to indicate the level of imbalance. The imbalance factors are 10, 50 and 100. We also test different openness by setting the last two or three classes to be unknown classes.

{\textbf{DomainNet}} \cite{domainnet} is a large-scale dataset that contains 6 domains with about 0.6 million images in 65 classes. Tan {\textit{et al.}} \cite{coal} resamples the dataset into 40 classes from 4 domains (Real (\textbf{R}), Clipart (\textbf{C}), Painting(\textbf{P}) and Sketch (\textbf{S})). As DomainNet exists salient heterogeneous
label shift, we set the last 20 classes in alphabetical order to be unknown samples. We utilize these four domains in turn as the source domain and target domain to build 12 tasks.

\subsection{Implementation Details}

We implement our framework in PyTorch \cite{torch}. We use ResNet-50 \cite{resnet} pretrained on ImageNet \cite{imagenet} as the backbone feature extractor in all experiments for a fair comparison. The last linear layer of ResNet-50 is replaced by a new weight matrix to construct ${\boldsymbol{W}}$. For a fair comparison, the shared hyperparameters of DANCE and OMEGA are the same. Specifically, we set $\tau$ in \cref{temp} as 0.05, $\eta_1$ in \cref{all} as 0.05, $\eta_2$ in \cref{all} as 0.1, $m$ in \cref{l_es} as 0.5, $Z = 0.5 \cdot K$, $r=0.15$ for all datasets. For Office-Home (RS-UT) and DomainNet, the batch-size is set to be 32 and the learning rate to be 0.01. For VisDA-C (RS-UT), the batch-size is set to be 128 for acceleration. According to \cite{lr}, when the batch-size becomes $N$ times the original size, the learning rate should be increased to $\sqrt{N}$ times the original size, so the learning rate on VisDA-C (RS-UT) is 0.02. We use stochastic gradient descent (SGD) optimizer with momentum 0.9 to optimize the models. The learning rate is decayed with the factor of $(1 + \gamma \frac{iter}{10000}) ^ {p}$ with $\gamma = 10$, $p=-0.75$. Please refer to our code for more details. 

We adopt the HOS score \cite{ros} as evaluation metric. As there exists class imbalance problem, we set the per-class average accuracy over the known classes as $\rm{OS^{*}}$ and the accuracy of the unknown samples as $\rm{UNK}$. HOS is the harmonic mean of $\rm{OS^{*}}$ and $\rm{UNK}$.
\begin{equation}
  {\rm{OS^{*}}} = \frac{1}{|\mathcal{C}_s|} \sum_{i=1}^{|\mathcal{C}_s|} \frac{|x: x \in \mathcal{D}_{i}^{t} \wedge \hat{y}(x) = i|}{|x: x \in \mathcal{D}_{i}^{t}|}
\end{equation}
\begin{equation}
  {\rm{UNK}} = \frac{|x: x \in \mathcal{D}_{|\mathcal{C}_s| + 1}^{t} \wedge \hat{y}(x) = |\mathcal{C}_s| + 1|}{|x: x \in \mathcal{D}_{|\mathcal{C}_s| + 1}^{t}|}
\end{equation}
\begin{equation}
  {\rm{HOS}} = 2 \times \frac{{\rm{OS^{*}}} \times {\rm{UNK}}}{{\rm{OS^{*}}} + {\rm{UNK}}}
\end{equation}

\subsection{Experimental Results}

\begin{figure*}[]
  \centering
  \subfloat[]{\includegraphics[width=2.2in, trim={45mm 15mm 15mm 45mm}, clip]{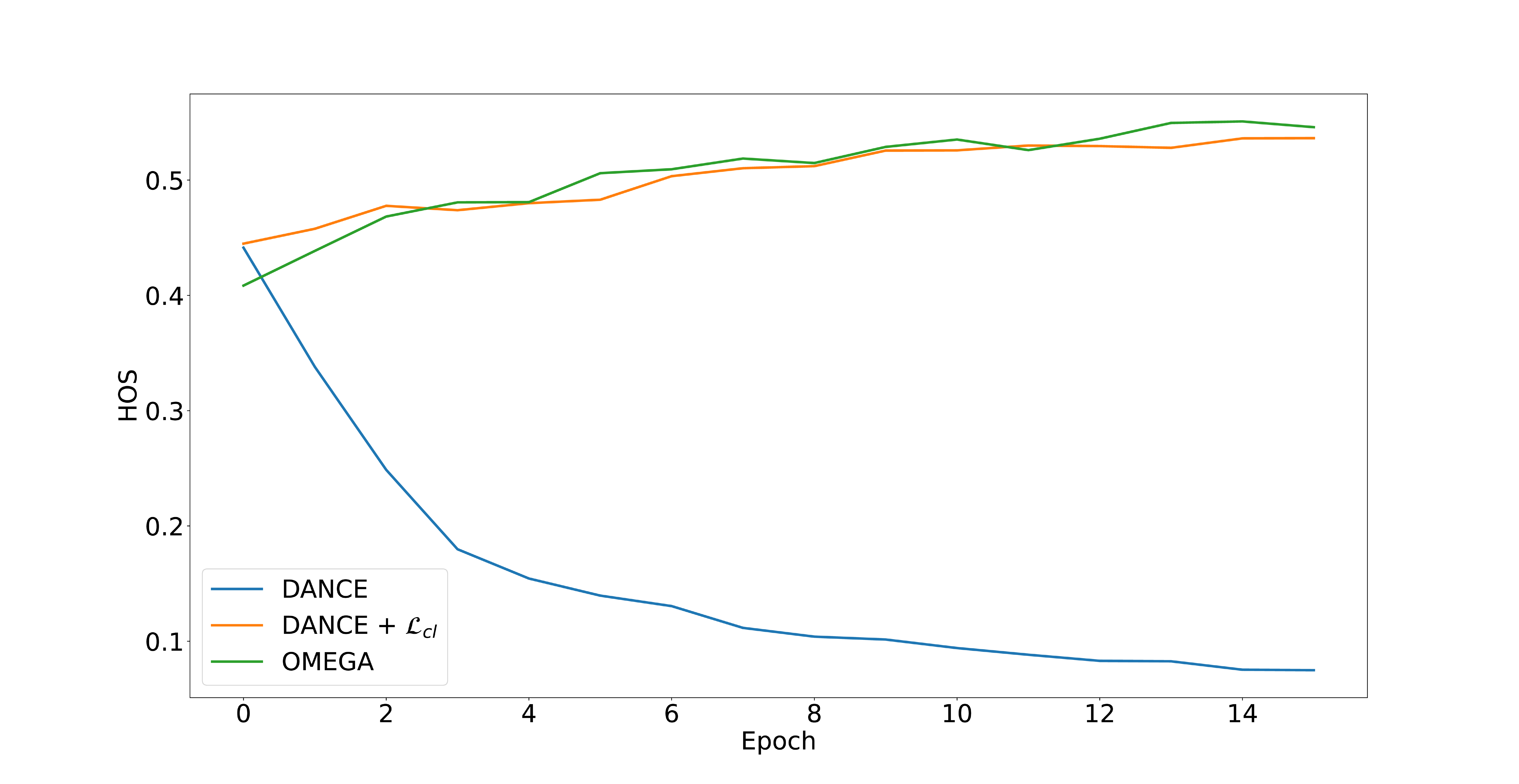}%
  \label{hos_fig}}
  \hfil
  \subfloat[]{\includegraphics[width=2.2in, trim={45mm 15mm 15mm 45mm}, clip]{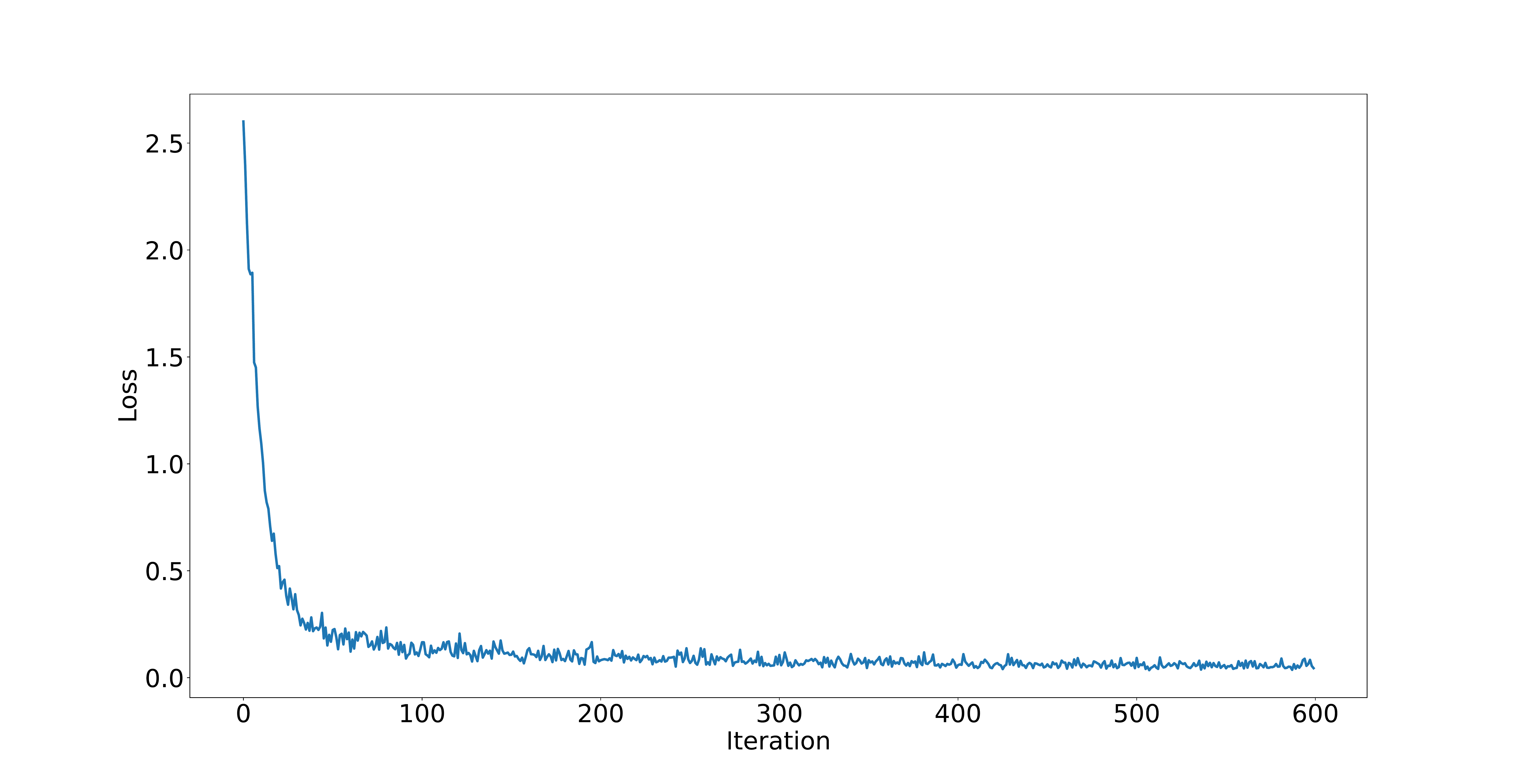}%
  \label{loss_fig}}
  \hfil
  \subfloat[]{\includegraphics[width=2.2in, trim={30mm 5mm 5mm 30mm}, clip]{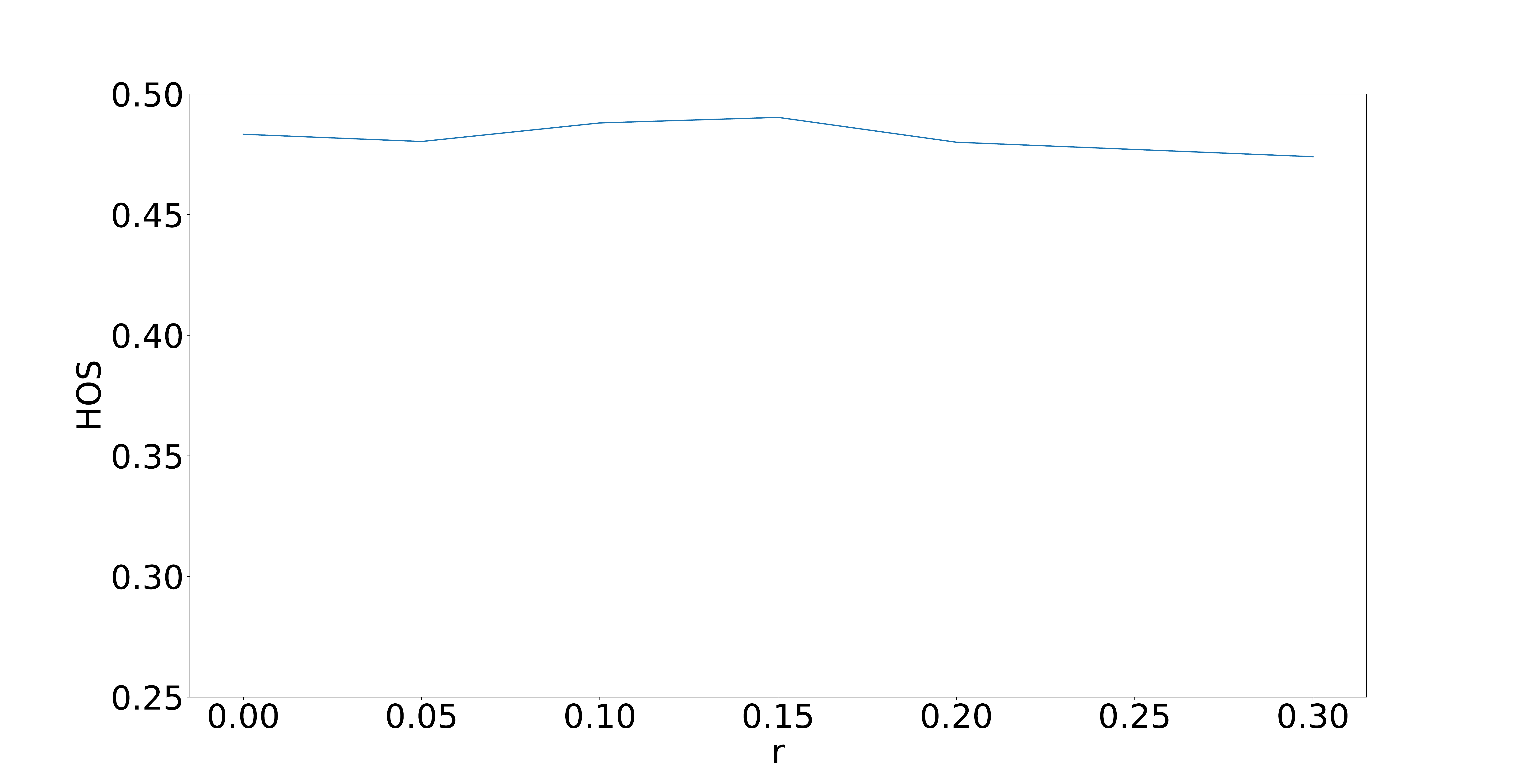}%
  \label{sen_fig}}
  \caption{Experiment Analysis. (a) HOS curve for ablation study. (b) Convergence curve of the total loss in OMEGA. (c) Sensitivity to the hyperparameter $r$.}
  \label{analysis}
\end{figure*}

\begin{figure*}[]
  \centering
  \subfloat[]{\includegraphics[width=2.2in, trim={100mm 45mm 80mm 40mm}, clip]{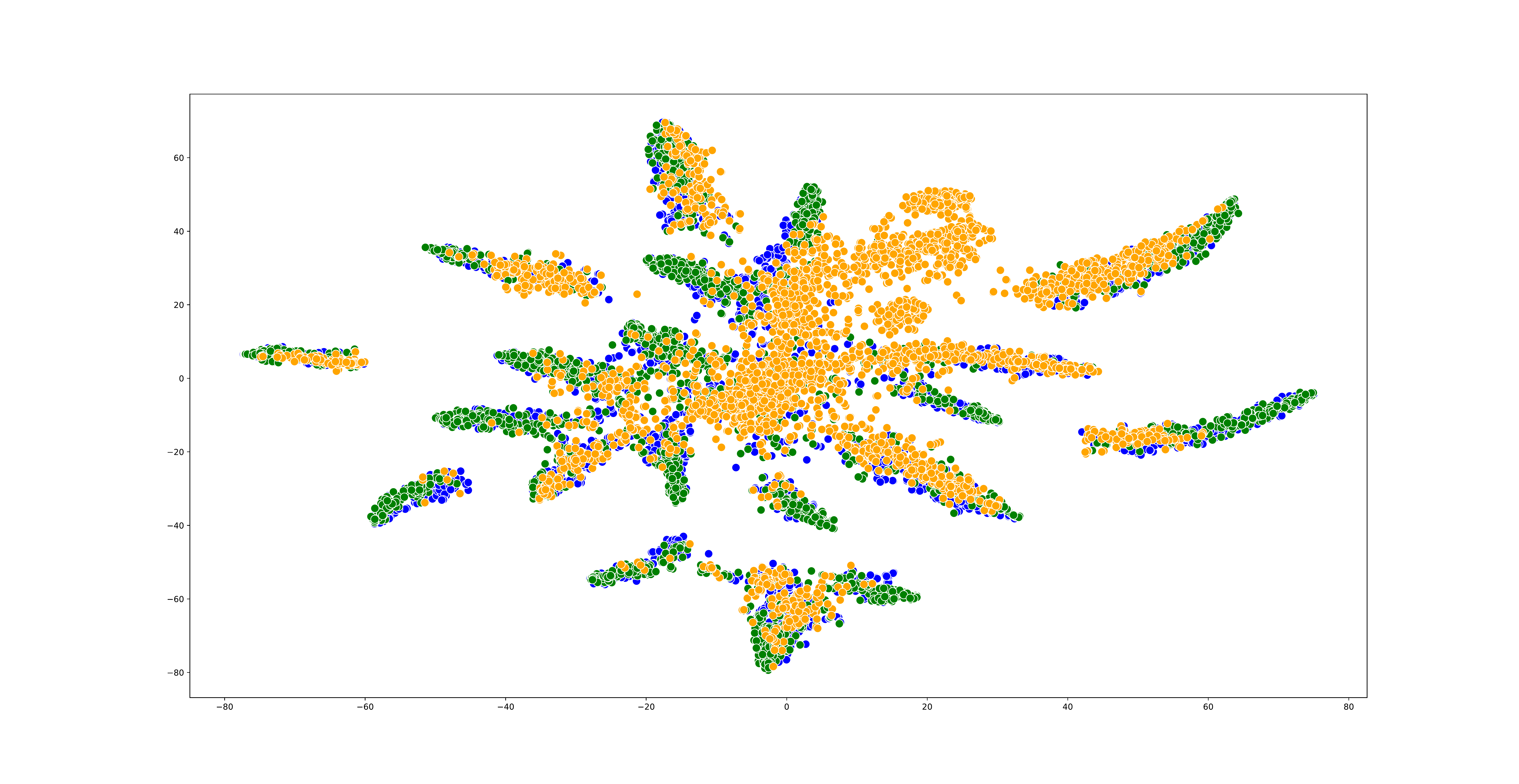}%
  \label{osbp_vis}}
  \hfil
  \subfloat[]{\includegraphics[width=2.2in, trim={100mm 45mm 80mm 40mm}, clip]{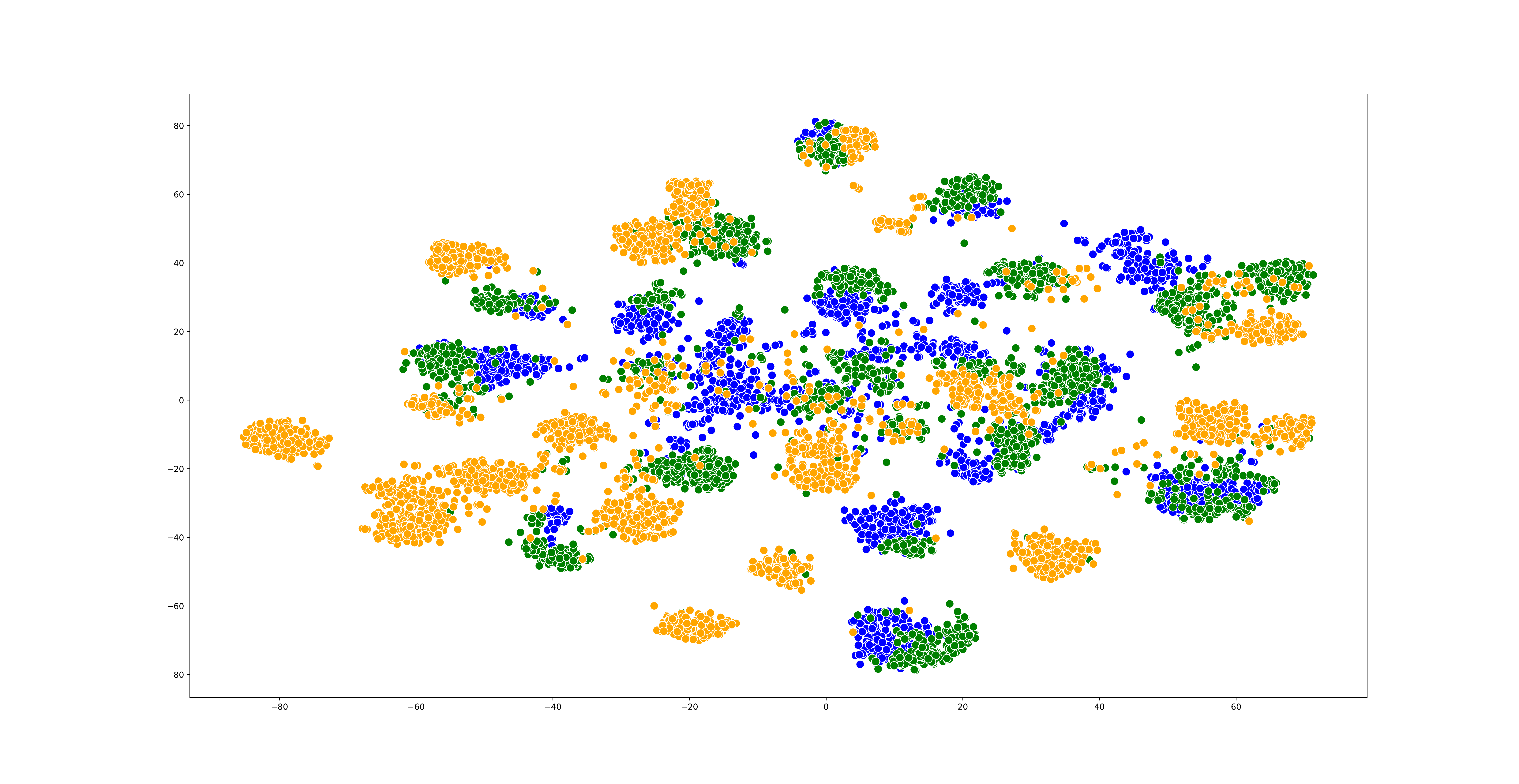}%
  \label{dance_vis}}
  \hfil
  \subfloat[]{\includegraphics[width=2.2in, trim={100mm 45mm 80mm 40mm}, clip]{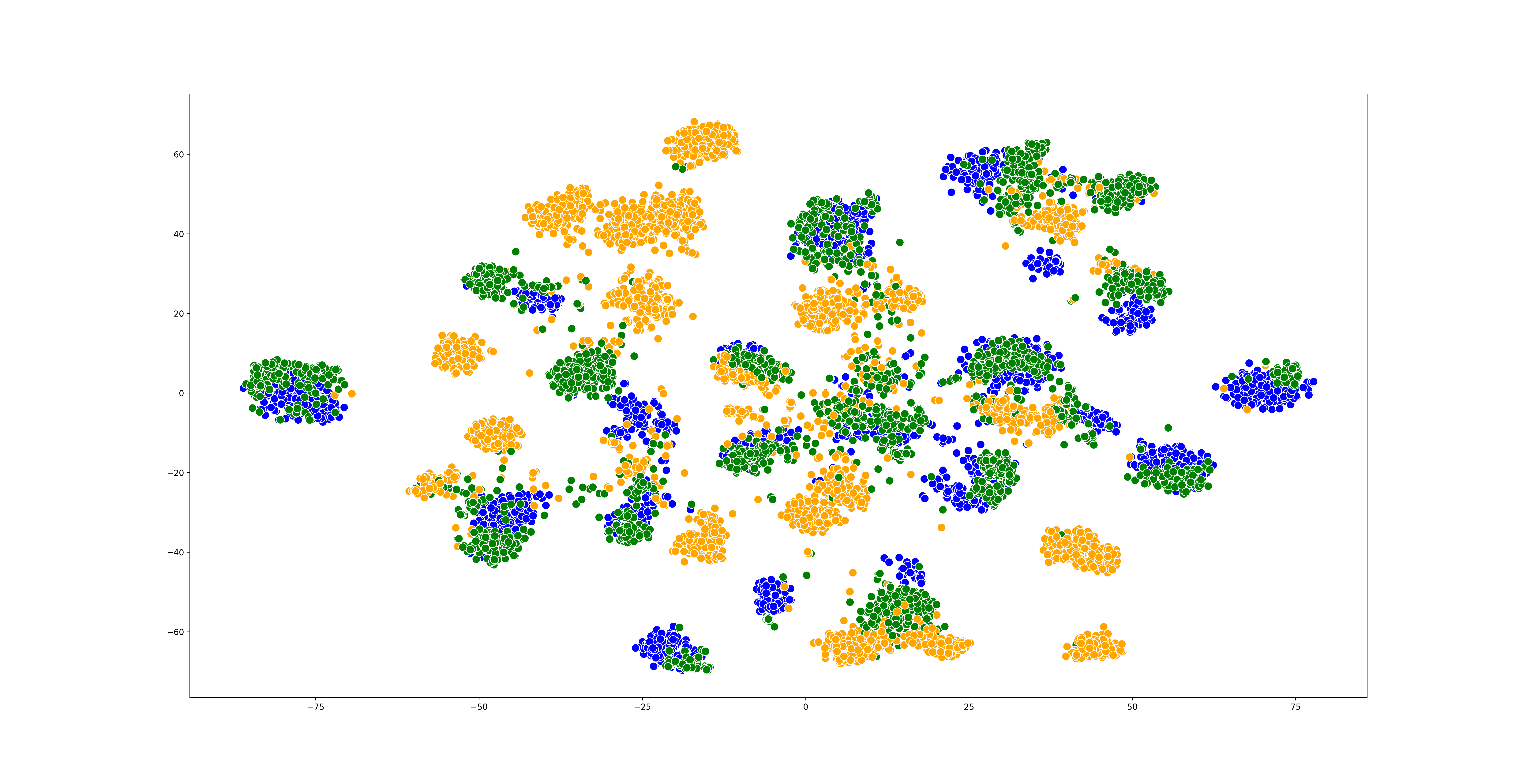}%
  \label{omega_vis}}
  \caption{Visualization of features using t-SNE. We take S$\rightarrow$R on DomainNet as an example. Blue, green and orange points represent source samples, target samples of the shared classes and unknown target samples. (a) OSBP. (b) DANCE. (C) OMEGA.}
  \label{vis}
\end{figure*}

The results of IOSDA on the 6 tasks on Office-Home (RS-UT), 12 tasks on DomainNet, and 6 tasks on VisDA-C (RS-Ut) are shown in \Cref{result_oh}-\Cref{result_dn}, respectively. We compare OMEGA with three state-of-the-art OSDA methods (STA \cite{sta}, OSBP \cite{osbp} and ROS \cite{ros}), four UniDA methods (UAN \cite{unida}, CMU \cite{cmu}, DANCE \cite{dance} and Ovanet \cite{ova}) and the OSLS method PULSE \cite{osls}. All results are reproduced by us with the official codes and hyperparameters.

On Office-Home dataset, our method achieves the best results on all tasks except for the Pr $\rightarrow$ Rw task and 2.73\% improvement over the optimal baseline method DANCE. On DomainNet dataset, our method achieves the best results on 11 out of 12 tasks and outperforms Ovanet by 3.59\%. On VisDA-C dataset, our method achieves the best results on 5 out of 6 tasks and improves the best baseline Ovanet with an advantage of 3.94\%. All 24 tasks validate that OMEGA is suitable for medium and large scale datasets with large domain shift.

From the results, we could observe that OMEGA outperforms all baseline methods by a large scale on all datasets. It can be seen that OMEGA could outperform OSDA methods by nearly 10\% in terms of HOS score. Common OSDA methods degrade severely on imbalanced datasets because they typically focus on aligning marginal distribution which may harm the performance on minority classes. As for UniDA methods, experiments show that state-of-the-art UniDA methods like DANCE and Ovanet tend to perform well on IOSDA setting because the ability of UniDA methods to detect source-private classes could also be used to ease the influence of class imbalance problem. PULSE is the first OSLS method that is close to our setting. However, experiments validate that PULSE tends to treat all samples as unknown on most of the tasks. This is because that the design of PULSE is based on precise mathematical derivation and PULSE relies heavily on the assumption $p(x|y) = q(x|y)$. Therefore, PULSE is not robust enough when there exists covariate shift.

To further demonstrate the robustness and effectiveness of OMEGA, we conduct experiments on OSDA and UniDA benchmarks
in the same settings with baseline methods. The results are shown in \Cref{result_osda}-\Cref{result_opda}. Note that the results are produced with no hyperparameters tuned. In OSDA setting, OMEGA could outperform OSBP and STA, achieving higher HOS scores than ROS on $7/12$ tasks and getting the highest average score. Considering that these methods are specifically designed for OSDA, we believe this result is satisfactory. The performance of the remaining UniDA methods is not as good as the OSDA methods because they are not tailored for OSDA problems. In OPDA setting, OMEGA could significantly outperform all baselines and outperform Ovanet by 3.7\%. This is due to the fact that OMEGA can handle class imbalance well, while OPDA can be seen as a special case of class imbalance. This result may provide a new perspective for dealing with the ordinary UniDA problem, {\textit{i.e.}}, solving UniDA by drawing on the approach for dealing with class imbalance problems.

From the experimental results, it's apparent that OMEGA could achieve good performance not only on IOSDA benchmarks, but also on OSDA and OPDA tasks with very competitive results. 

\subsection{Analysis}
{\textbf{Feature Visualization}}. We visualize the feature distributions of S$\rightarrow$R task on DomainNet using t-SNE \cite{tsne}. As shown in \cref{vis}, compared to OSBP and DANCE, the proposed OMEGA form discriminative clusters in both source domain and target domain. The known samples of the target domain are successfully aligned with the source domain, and the unknown samples are kept away from known ones. However, OSBP is not able to distinguish samples of different categories, and samples of unknown and known categories are randomly mixed together under extreme class imbalance. DANCE is able to distinguish unknown from known, but does not align the samples in the source and target domain well. With moving-threshold estimation and unknown-aware target clustering loss, OMEGA is able to generate thresholds for each sample and align the source and target domains by forming tight clusters.

{\textbf{Ablation Study}}. We conduct experiments on VisDA-C (RS-UT) to validate the effectiveness of our moving-threshold estimation (ME) and unknown-aware target clustering loss ($\mathcal{L}_{cl}$). The results are shown in \Cref{ablation} and \cref{hos_fig}. The moving-threshold estimation scheme leads to considerable improvements in all 6 tasks and improves the baseline DANCE \cite{dance} on average 3.05\%. Furthermore, $\mathcal{L}_{cl}$ could not only improve the accuracy of the baseline method by a large scale but also improve the stability of it, which is shown in \cref{hos_fig}. The performance of DANCE will degrade severely after a few iterations. However, OMEGA could achieve higher HOS score while maintaining high stability during training, which benefits from the discriminative clusters formed by $\mathcal{L}_{cl}$. 

{\textbf{Label Shift Analysis}}. As shown in \Cref{result_visda}, the performances of common open set domain adaptation and universal domain adaptation methods seriously drop under class imbalance. When there is no label shift, the HOS metric of experiments on Office-Home dataset using ResNet-50 could reach 70\% \cite{ros}, but it is less than 60\% in our setting. To verify the robustness of these methods, we conduct experiments to test different imbalance factors on the VisDA-C dataset. When there exists different degrees of label shift, the OSDA methods tend to perform similarly on different tasks. However, we notice that the performance of certain methods is not stable as $\omega$ changes, but OMEGA remains high score during all tasks, which verifies the robustness and consistent performance of our method.

{\textbf{Openness Analysis}}. We define the {\textit{openness}} $\mathbb{O} = \frac{N_{unk}}{N_t}$ where ${N_{unk}}$ is the number of unknown samples. Following previous OSDA methods \cite{ros}, we set $\mathbb{O}$ near 0.5 for most of the experiments. However, as we treat the number of unknown classes on each dataset as a fixed value, the {\textit{openness}} $\mathbb{O}$ is not constant for each task because the number of unknown samples may change. For example, the openness is 0.433 in the P$\rightarrow$C task of DomainNet, and it becomes 0.505 in the P$\rightarrow$S task. Apart from this, in order to verify the robustness of OMEGA under different degrees of $\mathbb{O}$, we further change the classes of unknown samples of VisDA-C dataset. In this way, the openness increases from 0.372 when $\omega = 10, K = 10$ to 0.720 when $\omega = 100, K = 9$. As shown in \Cref{result_visda}, some methods like OSBP \cite{osbp} are not robust to the openness, but OMEGA maintain satisfactory performance under different degrees of openness.

{\textbf{Hyperparameter Sensitivity}}. We conduct experiments based on VisDA-C ($\omega = 10, K = 10$) to evaluate the sensitivity of $r$ in moving-threshold estimation. We tune $r$ from 0 to 0.3 to get different HOS score. As shown in \cref{sen_fig}., the HOS has low sensitivity to $r$, showing the reliability of OMEGA. When $r = 0$, the moving-threshold estimation has no effect. The accuracy improves as $r$ increases, but if $r$ is too big, the HOS will drop a little, showing that a few samples are close to the decision boundary.

We also perform experiments to verify the other hyperparameters introduced by OMEGA. The results are shown in \Cref{zsens}. It's evident that OMEGA is not sensitive to $Z$ and $\eta_2$ even when $\eta_2 = 10\eta_1$. Our choices of HPs are empirically chosen without carefully tuned.

Furthermore, \Cref{result_osda} and \Cref{result_opda} show that OMEGA is robust enough because it could perform well on different tasks with the same hyperparameters.

{\textbf{Convergence Analysis}}. In order to verify the convergence tendency of our method, we illustrate the curve of the total loss of our method on the VisDA-C task when $\omega=10, K=10$ in \cref{loss_fig}. It can be seen that our method can effectively reduce the loss and rapidly converge to a small value in the first 500 iterations, which proves that the training process is smooth and convergent.

\section{Conclusion}
\label{con}
In this article, we introduce the IOSDA problem, a novel and practical UDA setting where the label shift exists on the basis of common OSDA and UniDA scenarios. Furthermore, to address this issue, we propose a novel framework called OMEGA (Open-set Moving-threshold Estimation and Gradual Alignment) model for IOSDA. By forming discriminative clusters using the proposed unknown-aware target clustering loss and generating specific thresholds by moving-threshold estimation for target samples, our model could effectively identify those known and unknown instances while gradually align the source and the target domain. Extensive experiments on three large datasets and on OSDA/OPDA settings verify the effectiveness and robustness of OMEGA.

For the future work, we believe our method could be easily extended to universal domain adaptation and could be used in common open set domain adaptation problems. Furthermore, as OMEGA is tailored for real-world scenarios, it could be easily applied to many industrial tasks, such like the machinery fault diagnosis \cite{machine} and the machine remaining useful life prediction \cite{rulp,wkk}. 

\bibliographystyle{ieeetran}
\bibliography{iosda.bib}

\begin{thebibliography}{10}
\providecommand{\url}[1]{#1}
\csname url@samestyle\endcsname
\providecommand{\newblock}{\relax}
\providecommand{\bibinfo}[2]{#2}
\providecommand{\BIBentrySTDinterwordspacing}{\spaceskip=0pt\relax}
\providecommand{\BIBentryALTinterwordstretchfactor}{4}
\providecommand{\BIBentryALTinterwordspacing}{\spaceskip=\fontdimen2\font plus
\BIBentryALTinterwordstretchfactor\fontdimen3\font minus
  \fontdimen4\font\relax}
\providecommand{\BIBforeignlanguage}[2]{{%
\expandafter\ifx\csname l@#1\endcsname\relax
\typeout{** WARNING: IEEEtran.bst: No hyphenation pattern has been}%
\typeout{** loaded for the language `#1'. Using the pattern for}%
\typeout{** the default language instead.}%
\else
\language=\csname l@#1\endcsname
\fi
#2}}
\providecommand{\BIBdecl}{\relax}
\BIBdecl

\bibitem{alexnet}
A.~Krizhevsky, I.~Sutskever, and G.~E. Hinton, ``Imagenet classification with
  deep convolutional neural networks,'' \emph{Communications of the ACM},
  vol.~60, no.~6, pp. 84--90, 2017.

\bibitem{resnet}
K.~He, X.~Zhang, S.~Ren, and J.~Sun, ``Deep residual learning for image
  recognition,'' in \emph{Proceedings of the IEEE conference on computer vision
  and pattern recognition}, 2016, pp. 770--778.

\bibitem{maskrcnn}
K.~He, G.~Gkioxari, P.~Doll{\'a}r, and R.~Girshick, ``Mask r-cnn,'' in
  \emph{Proceedings of the IEEE international conference on computer vision},
  2017, pp. 2961--2969.

\bibitem{mma}
A.~Aslam and E.~Curry, ``Reducing response time for multimedia event processing
  using domain adaptation,'' in \emph{Proceedings of the 2020 International
  Conference on Multimedia Retrieval}, 2020, pp. 261--265.

\bibitem{mmda1}
Y.~Peng and J.~Chi, ``Unsupervised cross-media retrieval using domain
  adaptation with scene graph,'' \emph{IEEE Transactions on Circuits and
  Systems for Video Technology}, vol.~30, no.~11, pp. 4368--4379, 2019.

\bibitem{mmda2}
R.~Wang, Z.~Wu, Z.~Weng, J.~Chen, G.-J. Qi, and Y.-G. Jiang, ``Cross-domain
  contrastive learning for unsupervised domain adaptation,'' \emph{IEEE
  Transactions on Multimedia}, 2022.

\bibitem{cdcl}
------, ``Cross-domain contrastive learning for unsupervised domain
  adaptation,'' \emph{IEEE Transactions on Multimedia}, 2022.

\bibitem{uda}
Y.~Ganin and V.~Lempitsky, ``Unsupervised domain adaptation by
  backpropagation,'' in \emph{International conference on machine
  learning}.\hskip 1em plus 0.5em minus 0.4em\relax PMLR, 2015, pp. 1180--1189.

\bibitem{udar}
M.~Long, H.~Zhu, J.~Wang, and M.~I. Jordan, ``Unsupervised domain adaptation
  with residual transfer networks,'' \emph{Advances in neural information
  processing systems}, vol.~29, 2016.

\bibitem{mmd}
E.~Tzeng, J.~Hoffman, N.~Zhang, K.~Saenko, and T.~Darrell, ``Deep domain
  confusion: Maximizing for domain invariance,'' \emph{arXiv preprint
  arXiv:1412.3474}, 2014.

\bibitem{wasser1}
J.~Shen, Y.~Qu, W.~Zhang, and Y.~Yu, ``Wasserstein distance guided
  representation learning for domain adaptation,'' in \emph{Proceedings of the
  AAAI Conference on Artificial Intelligence}, vol.~32, no.~1, 2018.

\bibitem{dis1}
M.~Li, Y.-M. Zhai, Y.-W. Luo, P.-F. Ge, and C.-X. Ren, ``Enhanced transport
  distance for unsupervised domain adaptation,'' in \emph{Proceedings of the
  IEEE/CVF conference on computer vision and pattern recognition}, 2020, pp.
  13\,936--13\,944.

\bibitem{adda}
E.~Tzeng, J.~Hoffman, K.~Saenko, and T.~Darrell, ``Adversarial discriminative
  domain adaptation,'' in \emph{Proceedings of the IEEE conference on computer
  vision and pattern recognition}, 2017, pp. 7167--7176.

\bibitem{adv2}
K.~Saito, K.~Watanabe, Y.~Ushiku, and T.~Harada, ``Maximum classifier
  discrepancy for unsupervised domain adaptation,'' in \emph{Proceedings of the
  IEEE conference on computer vision and pattern recognition}, 2018, pp.
  3723--3732.

\bibitem{unif}
W.~Zhang, X.~Li, H.~Ma, Z.~Luo, and X.~Li, ``Universal domain adaptation in
  fault diagnostics with hybrid weighted deep adversarial learning,''
  \emph{IEEE Transactions on Industrial Informatics}, vol.~17, no.~12, pp.
  7957--7967, 2021.

\bibitem{osda}
P.~Panareda~Busto and J.~Gall, ``Open set domain adaptation,'' in
  \emph{Proceedings of the IEEE international conference on computer vision},
  2017, pp. 754--763.

\bibitem{osbp}
K.~Saito, S.~Yamamoto, Y.~Ushiku, and T.~Harada, ``Open set domain adaptation
  by backpropagation,'' in \emph{Proceedings of the European Conference on
  Computer Vision (ECCV)}, 2018, pp. 153--168.

\bibitem{bbse}
Z.~Lipton, Y.-X. Wang, and A.~Smola, ``Detecting and correcting for label shift
  with black box predictors,'' in \emph{International conference on machine
  learning}.\hskip 1em plus 0.5em minus 0.4em\relax PMLR, 2018, pp. 3122--3130.

\bibitem{pareto}
W.~J. Reed, ``The pareto, zipf and other power laws,'' \emph{Economics
  letters}, vol.~74, no.~1, pp. 15--19, 2001.

\bibitem{logit}
A.~K. Menon, S.~Jayasumana, A.~S. Rawat, H.~Jain, A.~Veit, and S.~Kumar,
  ``Long-tail learning via logit adjustment,'' in \emph{International
  Conference on Learning Representations}.

\bibitem{facelt}
Y.~Wen, K.~Zhang, Z.~Li, and Y.~Qiao, ``A comprehensive study on center loss
  for deep face recognition,'' \emph{International Journal of Computer Vision},
  vol. 127, no.~6, pp. 668--683, 2019.

\bibitem{face1}
D.~Cao, X.~Zhu, X.~Huang, J.~Guo, and Z.~Lei, ``Domain balancing: Face
  recognition on long-tailed domains,'' in \emph{Proceedings of the IEEE/CVF
  Conference on Computer Vision and Pattern Recognition}, 2020, pp. 5671--5679.

\bibitem{coal}
S.~Tan, X.~Peng, and K.~Saenko, ``Class-imbalanced domain adaptation: An
  empirical odyssey,'' in \emph{European Conference on Computer Vision}.\hskip
  1em plus 0.5em minus 0.4em\relax Springer, 2020, pp. 585--602.

\bibitem{unida}
K.~You, M.~Long, Z.~Cao, J.~Wang, and M.~I. Jordan, ``Universal domain
  adaptation,'' in \emph{Proceedings of the IEEE/CVF conference on computer
  vision and pattern recognition}, 2019, pp. 2720--2729.

\bibitem{neg_trans}
H.~Zhao, R.~T. Des~Combes, K.~Zhang, and G.~Gordon, ``On learning invariant
  representations for domain adaptation,'' in \emph{International Conference on
  Machine Learning}.\hskip 1em plus 0.5em minus 0.4em\relax PMLR, 2019, pp.
  7523--7532.

\bibitem{dance}
K.~Saito, D.~Kim, S.~Sclaroff, and K.~Saenko, ``Universal domain adaptation
  through self supervision,'' \emph{Advances in neural information processing
  systems}, vol.~33, pp. 16\,282--16\,292, 2020.

\bibitem{dsbb}
W.-G. Chang, T.~You, S.~Seo, S.~Kwak, and B.~Han, ``Domain-specific batch
  normalization for unsupervised domain adaptation,'' in \emph{Proceedings of
  the IEEE/CVF conference on Computer Vision and Pattern Recognition}, 2019,
  pp. 7354--7362.

\bibitem{yq}
K.~Weiss, T.~M. Khoshgoftaar, and D.~Wang, ``A survey of transfer learning,''
  \emph{Journal of Big data}, vol.~3, no.~1, pp. 1--40, 2016.

\bibitem{mmd1}
M.~Long, Y.~Cao, J.~Wang, and M.~Jordan, ``Learning transferable features with
  deep adaptation networks,'' in \emph{International conference on machine
  learning}.\hskip 1em plus 0.5em minus 0.4em\relax PMLR, 2015, pp. 97--105.

\bibitem{mmd2}
Z.~Wang, B.~Du, and Y.~Guo, ``Domain adaptation with neural embedding
  matching,'' \emph{IEEE transactions on neural networks and learning systems},
  vol.~31, no.~7, pp. 2387--2397, 2019.

\bibitem{wasserstein}
G.~Peyr{\'e}, M.~Cuturi \emph{et~al.}, ``Computational optimal transport: With
  applications to data science,'' \emph{Foundations and Trends{\textregistered}
  in Machine Learning}, vol.~11, no. 5-6, pp. 355--607, 2019.

\bibitem{gan}
I.~Goodfellow, J.~Pouget-Abadie, M.~Mirza, B.~Xu, D.~Warde-Farley, S.~Ozair,
  A.~Courville, and Y.~Bengio, ``Generative adversarial networks,''
  \emph{Communications of the ACM}, vol.~63, no.~11, pp. 139--144, 2020.

\bibitem{adv1}
Y.~Ganin and V.~Lempitsky, ``Unsupervised domain adaptation by
  backpropagation,'' in \emph{International conference on machine
  learning}.\hskip 1em plus 0.5em minus 0.4em\relax PMLR, 2015, pp. 1180--1189.

\bibitem{adv3}
J.~Li, E.~Chen, Z.~Ding, L.~Zhu, K.~Lu, and Z.~Huang, ``Cycle-consistent
  conditional adversarial transfer networks,'' in \emph{Proceedings of the 27th
  ACM international conference on multimedia}, 2019, pp. 747--755.

\bibitem{ros}
S.~Bucci, M.~R. Loghmani, and T.~Tommasi, ``On the effectiveness of image
  rotation for open set domain adaptation,'' in \emph{European Conference on
  Computer Vision}.\hskip 1em plus 0.5em minus 0.4em\relax Springer, 2020, pp.
  422--438.

\bibitem{psdc}
Z.~Liu, G.~Chen, Z.~Li, Y.~Kang, S.~Qu, and C.~Jiang, ``Psdc: A prototype-based
  shared-dummy classifier model for open-set domain adaptation,'' \emph{IEEE
  Transactions on Cybernetics}, 2022.

\bibitem{ova}
K.~Saito and K.~Saenko, ``Ovanet: One-vs-all network for universal domain
  adaptation,'' in \emph{Proceedings of the IEEE/CVF International Conference
  on Computer Vision}, 2021, pp. 9000--9009.

\bibitem{yjf}
J.~Yang, J.~Yang, S.~Wang, S.~Cao, H.~Zou, and L.~Xie, ``Advancing imbalanced
  domain adaptation: Cluster-level discrepancy minimization with a
  comprehensive benchmark,'' \emph{IEEE Transactions on Cybernetics}, 2021.

\bibitem{partial}
Z.~Cao, K.~You, M.~Long, J.~Wang, and Q.~Yang, ``Learning to transfer examples
  for partial domain adaptation,'' in \emph{Proceedings of the IEEE/CVF
  conference on computer vision and pattern recognition}, 2019, pp. 2985--2994.

\bibitem{pda1}
Z.~Cao, L.~Ma, M.~Long, and J.~Wang, ``Partial adversarial domain adaptation,''
  in \emph{Proceedings of the European conference on computer vision (ECCV)},
  2018, pp. 135--150.

\bibitem{osls}
S.~Garg, S.~Balakrishnan, and Z.~C. Lipton, ``Domain adaptation under open set
  label shift,'' in \emph{Advances in Neural Information Processing Systems}.

\bibitem{minimax}
K.~Saito, D.~Kim, S.~Sclaroff, T.~Darrell, and K.~Saenko, ``Semi-supervised
  domain adaptation via minimax entropy,'' in \emph{Proceedings of the IEEE/CVF
  International Conference on Computer Vision}, 2019, pp. 8050--8058.

\bibitem{proto_uda}
K.~Tanwisuth, X.~Fan, H.~Zheng, S.~Zhang, H.~Zhang, B.~Chen, and M.~Zhou, ``A
  prototype-oriented framework for unsupervised domain adaptation,''
  \emph{Advances in Neural Information Processing Systems}, vol.~34, pp.
  17\,194--17\,208, 2021.

\bibitem{proto1}
X.~Yue, Z.~Zheng, S.~Zhang, Y.~Gao, T.~Darrell, K.~Keutzer, and A.~S.
  Vincentelli, ``Prototypical cross-domain self-supervised learning for
  few-shot unsupervised domain adaptation,'' in \emph{Proceedings of the
  IEEE/CVF Conference on Computer Vision and Pattern Recognition}, 2021, pp.
  13\,834--13\,844.

\bibitem{tao}
G.~Hinton, O.~Vinyals, J.~Dean \emph{et~al.}, ``Distilling the knowledge in a
  neural network,'' \emph{arXiv preprint arXiv:1503.02531}, vol.~2, no.~7,
  2015.

\bibitem{lmle}
C.~Huang, Y.~Li, C.~C. Loy, and X.~Tang, ``Learning deep representation for
  imbalanced classification,'' in \emph{Proceedings of the IEEE Conference on
  Computer Vision and Pattern Recognition (CVPR)}, June 2016.

\bibitem{isfda}
X.~Li, J.~Li, L.~Zhu, G.~Wang, and Z.~Huang, ``Imbalanced source-free domain
  adaptation,'' in \emph{Proceedings of the 29th ACM International Conference
  on Multimedia}, 2021, pp. 3330--3339.

\bibitem{kcl}
B.~Kang, Y.~Li, S.~Xie, Z.~Yuan, and J.~Feng, ``Exploring balanced feature
  spaces for representation learning,'' in \emph{International Conference on
  Learning Representations}, 2020.

\bibitem{sta}
H.~Liu, Z.~Cao, M.~Long, J.~Wang, and Q.~Yang, ``Separate to adapt: Open set
  domain adaptation via progressive separation,'' in \emph{Proceedings of the
  IEEE/CVF Conference on Computer Vision and Pattern Recognition}, 2019, pp.
  2927--2936.

\bibitem{cmu}
B.~Fu, Z.~Cao, M.~Long, and J.~Wang, ``Learning to detect open classes for
  universal domain adaptation,'' in \emph{European Conference on Computer
  Vision}.\hskip 1em plus 0.5em minus 0.4em\relax Springer, 2020, pp. 567--583.

\bibitem{oh}
H.~Venkateswara, J.~Eusebio, S.~Chakraborty, and S.~Panchanathan, ``Deep
  hashing network for unsupervised domain adaptation,'' in \emph{Proceedings of
  the IEEE conference on computer vision and pattern recognition}, 2017, pp.
  5018--5027.

\bibitem{visda}
X.~Peng, B.~Usman, N.~Kaushik, D.~Wang, J.~Hoffman, and K.~Saenko, ``Visda: A
  synthetic-to-real benchmark for visual domain adaptation,'' in
  \emph{Proceedings of the IEEE Conference on Computer Vision and Pattern
  Recognition Workshops}, 2018, pp. 2021--2026.

\bibitem{domainnet}
X.~Peng, Q.~Bai, X.~Xia, Z.~Huang, K.~Saenko, and B.~Wang, ``Moment matching
  for multi-source domain adaptation,'' in \emph{Proceedings of the IEEE/CVF
  international conference on computer vision}, 2019, pp. 1406--1415.

\bibitem{torch}
A.~Paszke, S.~Gross, F.~Massa, A.~Lerer, J.~Bradbury, G.~Chanan, T.~Killeen,
  Z.~Lin, N.~Gimelshein, L.~Antiga \emph{et~al.}, ``Pytorch: An imperative
  style, high-performance deep learning library,'' \emph{Advances in neural
  information processing systems}, vol.~32, 2019.

\bibitem{imagenet}
J.~Deng, W.~Dong, R.~Socher, L.-J. Li, K.~Li, and L.~Fei-Fei, ``Imagenet: A
  large-scale hierarchical image database,'' in \emph{2009 IEEE conference on
  computer vision and pattern recognition}.\hskip 1em plus 0.5em minus
  0.4em\relax Ieee, 2009, pp. 248--255.

\bibitem{lr}
A.~Krizhevsky, ``One weird trick for parallelizing convolutional neural
  networks,'' \emph{arXiv preprint arXiv:1404.5997}, 2014.

\bibitem{tsne}
L.~Van~der Maaten and G.~Hinton, ``Visualizing data using t-sne.''
  \emph{Journal of machine learning research}, vol.~9, no.~11, 2008.

\bibitem{machine}
W.~Zhang, X.~Li, H.~Ma, Z.~Luo, and X.~Li, ``Open-set domain adaptation in
  machinery fault diagnostics using instance-level weighted adversarial
  learning,'' \emph{IEEE Transactions on Industrial Informatics}, vol.~17,
  no.~11, pp. 7445--7455, 2021.

\bibitem{rulp}
M.~Ragab, Z.~Chen, M.~Wu, C.~S. Foo, C.~K. Kwoh, R.~Yan, and X.~Li,
  ``Contrastive adversarial domain adaptation for machine remaining useful life
  prediction,'' \emph{IEEE Transactions on Industrial Informatics}, vol.~17,
  no.~8, pp. 5239--5249, 2020.

\bibitem{wkk}
K.~Wu, J.~Li, L.~Zuo, K.~Lu, and H.~T. Shen, ``Weighted adversarial domain
  adaptation for machine remaining useful life prediction,'' \emph{IEEE
  Transactions on Instrumentation and Measurement}, vol.~71, pp. 1--11, 2022.

\end{thebibliography}

\end{document}